\def\year{2022}\relax
\documentclass[letterpaper]{article} 
\usepackage{aaai22}  
\usepackage{times}  
\usepackage{helvet}  
\usepackage{courier}  
\usepackage[hyphens]{url}  
\usepackage{graphicx} 
\usepackage{natbib}  
\usepackage{caption} 
\DeclareCaptionStyle{ruled}{labelfont=normalfont,labelsep=colon,strut=off} 
\usepackage{graphicx}
\usepackage{algorithm}
\usepackage{algorithmic}
\usepackage{amsmath}
\usepackage{amsfonts}

\usepackage{hyperref}
\hypersetup{
    colorlinks=true,
    citecolor=black,
    urlcolor=magenta,
}

\usepackage{newfloat}
\usepackage{listings}
\floatstyle{ruled}
\newfloat{listing}{tb}{lst}{}
\floatname{listing}{Listing}

\title{Learning from the Tangram to Solve Mini Visual Tasks}
\author{
    Yizhou Zhao\textsuperscript{\rm 1},
    Liang Qiu\textsuperscript{\rm 1},
    Pan Lu\textsuperscript{\rm 1},
    Feng Shi\textsuperscript{\rm 1},
    Tian Han\textsuperscript{\rm 2},
    Song-Chun Zhu\textsuperscript{\rm 1}
}
\affiliations{
    \textsuperscript{\rm 1}UCLA Center for Vision, Cognition, Learning, and Autonomy \\
    \textsuperscript{\rm 2} Stevens Institute of Technology \\
    yizhouzhao@g.ucla.edu
}

\usepackage{bibentry}

\begin{document}

\maketitle

\begin{abstract}
Current pre-training methods in computer vision focus on natural images in the daily-life context. However, abstract diagrams such as icons and symbols are common and important in the real world. This work is inspired by Tangram, a game that requires replicating an abstract pattern from seven dissected shapes. By recording human experience in solving tangram puzzles, we present the Tangram dataset and show that a pre-trained neural model on the Tangram helps solve some mini visual tasks based on low-resolution vision. Extensive experiments demonstrate that our proposed method generates intelligent solutions for aesthetic tasks such as folding clothes and evaluating room layouts. The pre-trained feature extractor can facilitate the convergence of few-shot learning tasks on human handwriting and improve the accuracy in identifying icons by their contours. The Tangram dataset is available at \url{https://github.com/yizhouzhao/Tangram}.
\end{abstract}

\section{Introduction}

As many vision tasks are relevant, one would expect a model, particularly pre-trained from one dataset, to assist a different challenge. Traditionally, supervised pre-training on image classification has been employed to help object detection~\cite{shinya2019understanding} and semantic parsing~\cite{orsic2019defense}. Moreover, popular unsupervised pre-training has recently produced remarkable results in visual tasks such as image classification~\cite{chen2020generative} and clustering~\cite{chakraborty2020g}. The common datasets to train basic models include PASCAL VOC~\cite{everingham2010pascal}, ImageNet~\cite{imagenet_cvpr09}, and COCO~\cite{lin2014microsoft}, all of which contain photographs.

\begin{figure*}[t]
\begin{center}
    \includegraphics[width=.95\linewidth]{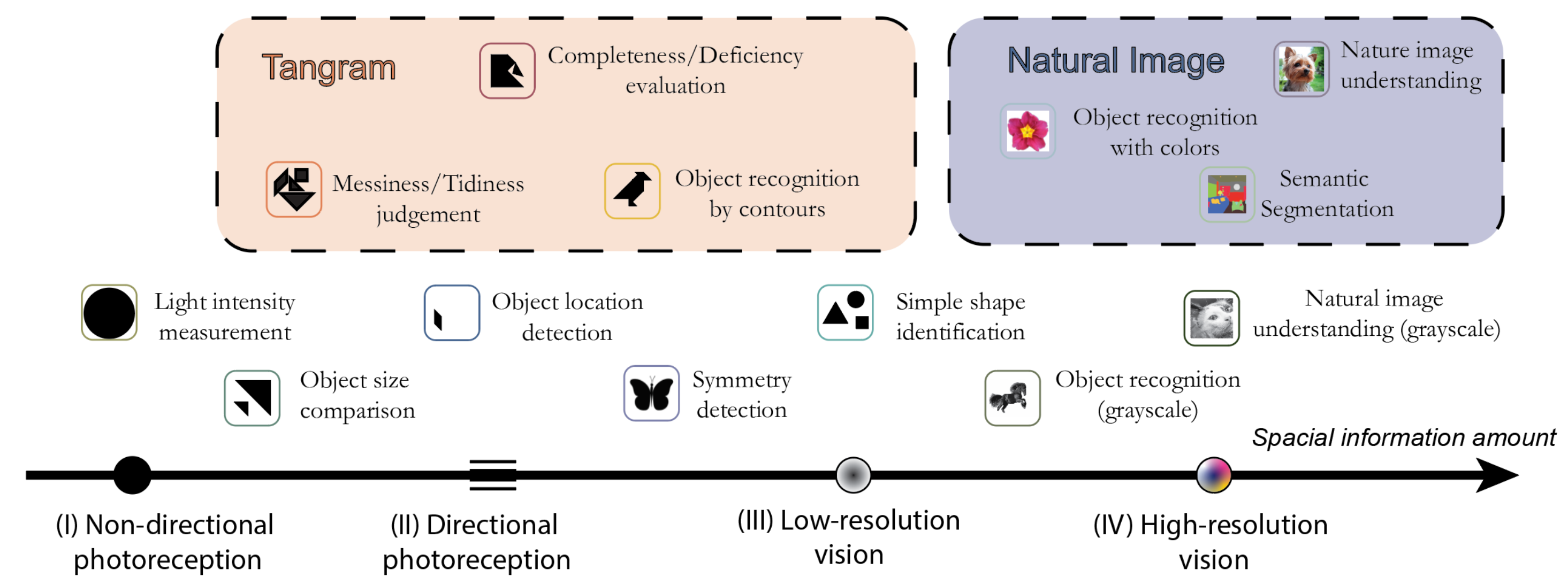}
\end{center}
  \caption{Visual perception tasks ranked by the amount of spatial information. In biology, visual perception tasks are divided into four levels based on the number of photoreceptors~\cite{land2012animal}. Our Tangram dataset relates to many low-resolution visual tasks, while current works usually focus on high-resolution natural images.}
\label{figure:eye}
\end{figure*}

It is natural to start the pre-training process from real-life images to solve daily vision tasks. However, one of the underlying limitations of current works is their focus on content from natural images. Besides natural images, abstract diagrams, such as texts, symbols, and signs, also carry rich visual semantics and account for a large part of the visual world. For instance, it is shown that emojis can express rich human sentiments \cite{felbo2017using}, and diagrams like icons can map the physical worlds into symbolic and aesthetic representations \cite{lagunas2019learning, madan2018synthetically, karamatsu2020iconify}. Furthermore, most of the tasks related to natural images can be accomplished by low-resolution vision~\cite{land2012animal} (see Figure~\ref{figure:eye}). Therefore, training an enormous backbone (e.g., a deep residual network~\cite{he2016deep}) to solve tasks related to abstract diagrams complicates the problem.

In this paper, we argue that we can solve the tasks related to abstract diagrams by learning from the process of replicating a tangram puzzle. The tangram, a dissection puzzle consisting of seven planar polygons (tans), is world-famous and has been used for many purposes, including art, design, and education. Although it only consists of seven tans, it can generate thousands of meaningful patterns such as animals, buildings, letters, and numbers. Solving a tangram puzzle associates with our cognitive and imaginative abilities.

We introduce the \textbf{Tangram}, a new dataset consisting of more than $10,000$ snapshots recording the steps to solve a total number of $388$ tangram puzzles. A neural model can be pre-trained from the Tangram to solve two groups of downstream tasks.

\begin{figure}[t]
\begin{center}
    \includegraphics[width=0.8\linewidth]{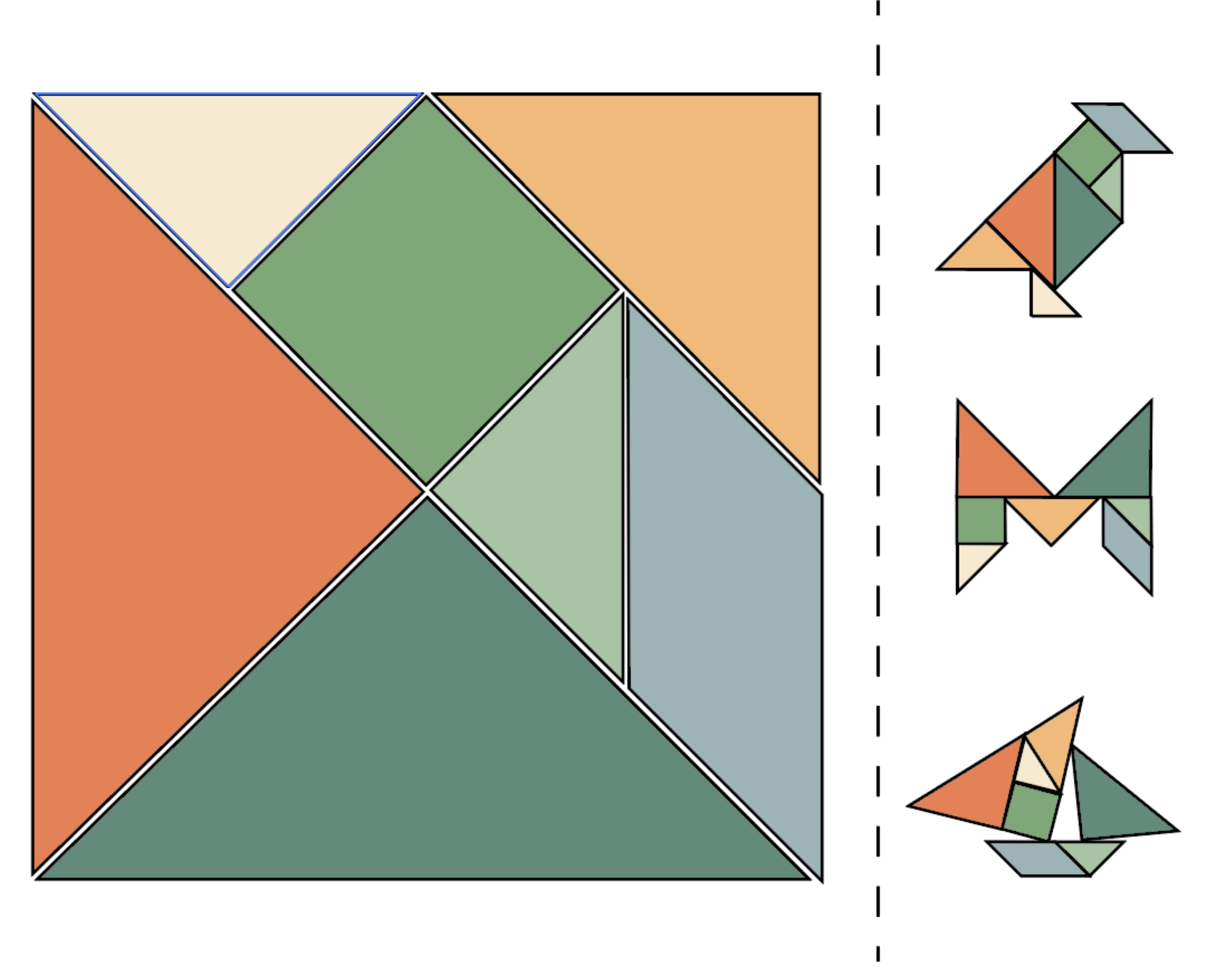}
\end{center}
  \caption{The left panel shows the square representation of the Tangram that consists of five triangles of three sizes, one parallelogram and one square. The right panel shows some tangram puzzles: a bird, the letter M and a sailboat.}
\label{fig:tangram}
\end{figure}

The first group is about aesthetics. We introduce two toy tasks: folding clothes and organizing furniture (room layouts). Tuning the pre-trained network from several expert samples can generate an aesthetic landscape that helps make aesthetic judgments. Experiments show that our method performs best when cooperating with max-entropy inverse reinforcement learning~\cite{ziebart2008maximum} and generative adversarial imitation learning~\cite{ho2016generative}.

The second group includes several recognition tasks. In the $N$-way-$K$-shot setting, we show that conducting pre-training on the Tangram improves the performance of recognizing the human handwriting, including Omniglot~\cite{lake2019omniglot} and Multi-digit MNIST~\cite{mulitdigitmnist}. This method also improves the performance of icon recognition from contours.

This paper makes three major contributions:
\begin{itemize}
    \item To our best knowledge, by introducing Tangram, we are the pioneers to suggest applying transfer learning from the human gaming experience to solve vision tasks.
    \item We demonstrate that pre-training from the Tangram can help solve both low-level aesthetics tasks and recognition tasks.
    \item We show that pretraining on the Tangram facilitates convergence in several few-shot learning tasks, and improves the performance of recognition in the setting of low-level vision.
\end{itemize}

\section{Related work}
An abundance of related work inspires our work, including pre-training in computer vision, rating image aesthetics with deep learning, and few-shot learning.

\subsection{Pre-training}
Pre-training methods can be either supervised or unsupervised. The supervised pre-training on ImageNet is conventional for object recognition, localization, and segmentation~\cite{he2019rethinking}. Inspired by the success of unsupervised pre-training in natural language processing, the community has gained much interest in studying unsupervised pre-training in computer vision, such as contrastive training~\cite{chen2020simple}, self-supervised training~\cite{jing2020self}. In many tasks, fine-tuning from a pre-trained model is faster than training from scratch. Pre-training can also help when high-quality labeled data is scarce.



\subsection{Image Aesthetics}
Image aesthetics assessment attempts to quantify an image's beauty. Image quality is influenced by numerous factors such as color \cite{nishiyama2011aesthetic}, lighting \cite{freeman2007complete}, texture \cite{ke2006design}, and image composition \cite{deng2017image}.  While subjective judgment by human eyes is the most reliable way to evaluate image quality, the beauty of an image can also be assessed by well-established photographic theories~\cite{zhai2020perceptual}. 
Recent research has shown that data-driven approaches can be more efficient, especially those that employ feature extraction by multi-column convolutional neural networks (CNNs)~\cite{lu2015rating, doshi2019image}.  Popular databases for image quality assessment (IQA)  are mainly collected as photos (natural images), such as the Photo.Net database~\cite{joshi2011aesthetics} and the CUHK-PhotoQuality database~\cite{luo2011content}. Some emerging databases consist of images from virtual contents such as screen content image quality database (SCIQ)~\cite{ni2017esim} and compressed Virtual reality image quality database (CVIQ)~\cite{sun2019mc360iqa}.


\subsection{Few-shot learning}
The main goal of few-shot learning is to learn new tasks with a few support examples while maintaining the ability to generalize.
Recently, there has been a growing interest in achieving the goal by learning prior knowledge from previous tasks, especially training feature extractors that can efficiently segregate novel classes~\cite{hu2020leveraging}. 

We apply our Tangram dataset to train the feature-extracting parts of optimization-based meta-learning algorithms such as MAML~\cite{finn2017model} and ANIL~\cite{raghu2019rapid}. Besides, since the Tangram only contains shapes and contours, we perform experiments on the few-shot learning tasks that are color-free and texture-free, for example, the Omniglot challenge~\cite{lake2019omniglot}.
\section{Pre-training from the Tangram}\label{sec:pre}
\subsection{Data collection}
To collect the process of solving puzzles from human experience, an interactive labeling tool is developed using the Unity game engine~\cite{haas2014history}. The labeling tool can record every step of moving, rotating, or flipping of one tan as a snapshot. Seven lab technicians spent weeks on completing a total number of $776$ solutions to $388$ unique puzzles, capturing more than $10,000$ snapshots.

\begin{figure}[!h]
    \centering
    \includegraphics[width=3in]{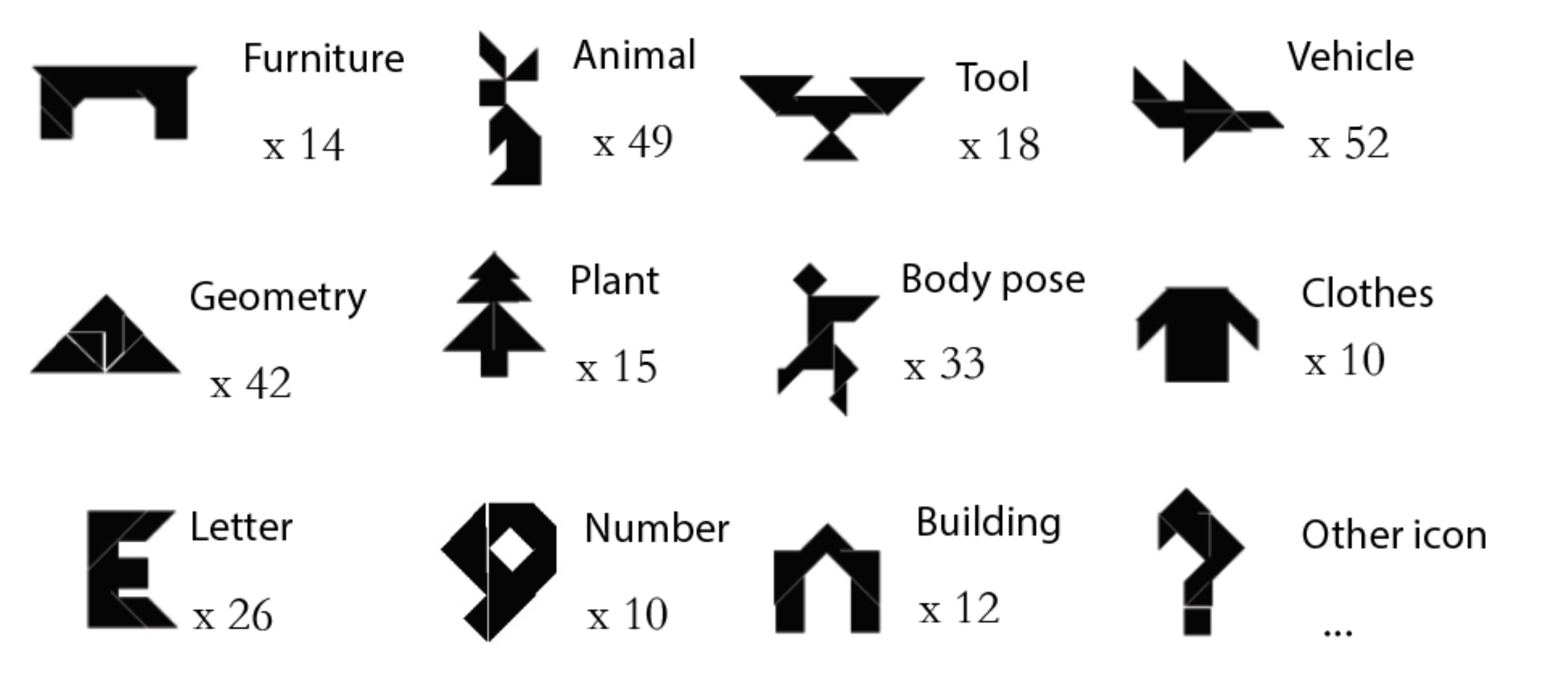}
    \caption{Collected examples of different categories in the Tangram dataset.}
    \label{figure:datacollection}
\end{figure}

Figure \ref{figure:datacollection} illustrates an overview of the puzzles types and their counts. The Tangram dataset consists of diverse tangram patterns including animals, plants, letters, numbers, buildings, human poses, and some everyday objects. It requires necessary perceptive recognition and elementary geometry skills to solve them. We will release the dataset to the public to encourage further study into abstract image understanding.

\subsection{Learning from puzzles}

Denote the order set $(I_1, I_2,..., I_{n_p})$ as the process to solve a tangram puzzle $P$, where each $I_i, i \in \{1,...,n_p\}$ is an image representing one step toward the solution, and $n_p$ is the total number of steps. Since a tangram pattern only has shapes and contours, $I_i$ is a binary image with size $H\times W$. 

What can we learn from the puzzles, and how can we use the solving steps? We argue that the Tangram reveals two pieces of information: 
\begin{itemize}
    \item The step-by-step solving process leads to more complete and tidy shape combinations, containing the perception of beauty.
    \item There is a connection between the pattern and the name of the object due to correspondence between the final completed pattern and a real-world object.
\end{itemize}
Therefore, we formulate two learning goals and assign two loss functions.

Let $f_\theta: \{0,1\}^{H \times W} \mapsto [0,1]$ be the function indicating the degree of completeness of step $I_i$. We define the \textbf{completeness contrast loss} (CCL) for the process $(I_{i})_{i=1}^{n_p}$ as 
\begin{align}\label{eq:ccl}
    \text{CCL}(I_1,...,I_p) &= (0 - f_{\theta}(I_{1}))^2 \nonumber \\ 
    & + \sum_{t=1}^{n_p-1} \Big(f_{\theta}(I_t) - f_{\theta}(I_{t+1})\Big)^2 \nonumber \nonumber \\ 
    & + (f_{\theta}(I_{n_p}) - 1)^2 .
\end{align}
By the Cauchy–Schwarz inequality, CCL reaches minimum value $\frac{1}{n_p + 1}$ when $f_\theta(I_i) = i/ (n_p + 1), i = 1, 2,...,n_p$. Minimizing CCL results in a right order for $(I_{i})_{i=1}^{n_p}$.

Let $g_\phi: \{0,1\}^{H \times W} \mapsto \mathbb{R}^{N_{\text{word}}}$ map the binary image to the word embedding $W_P$ of a pattern $P$, where $N_{\text{word}}$ is the dimension of the embedding space. The \textbf{puzzle meaning loss} (PML) for the final step $I_{n_p}$ is defined as
\begin{align}
    \text{PML}(I_{n_p}) = |g_\phi(I_{n_p}) - W_P|^2 .
\end{align}

\begin{figure}[t]
\begin{center}
    \includegraphics[width=0.8\linewidth]{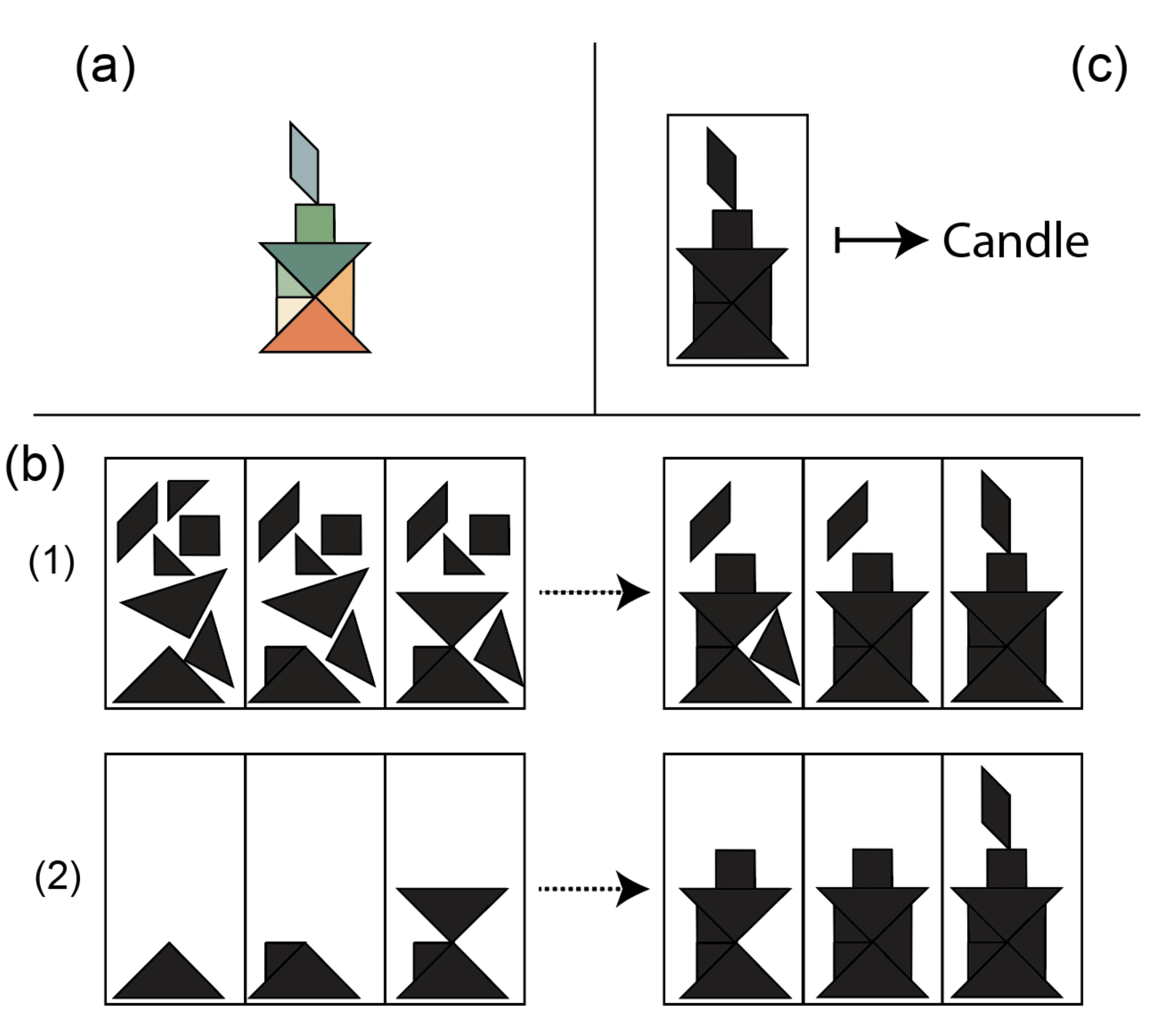}
\end{center}
  \caption{ (a) The expected solution of a tangram puzzle. (b) The process of solving the puzzle with its two variants. (c) The final completed puzzle image and the meaning of the item.}
\label{fig:loss}
\end{figure}

Figure \ref{fig:loss} depicts an implementation of the two loss functions described above. Panel (b) demonstrates two variants of the puzzle-solving processes. The first variant traces all tans, recording progression from disorganization to neatness; the second variant traces only the final state of moved tans and represents a progression from fragmentation to completeness.

To train the functions $f_\theta$ and $g_\phi$, we use a simple convolutional neural network with only four $3\times 3$ convolutional layers. Each image is resized into $28\times 28$. We apply the $50$-dimension GloVe embedding \cite{pennington2014glove} for pattern names, and we assign $80\%$ of the weight on CCL and $20\%$ on PML.
The feature extraction part of the network is transferred to achieve other challenges.

\section{Experiments}
 We define \textbf{mini visual tasks} as the vision tasks that only require learning from low-resolution binary images. We divide mini visual tasks into two categories: aesthetic tasks and recognition tasks. We choose folding clothes and generating room layouts (organizing furniture) as representatives for the first category, identifying human hand-writings and recognizing icons for the second. 
 
 \subsection{Folding clothes} \label{sec:folding_clothes}
Folding clothes is a classic task in robotics that has received heated discussion among various works. Prevalent methods include grounding human demonstration from videos~\cite{yang2015robot}, employing random decision forests and probabilistic planning~\cite{doumanoglou2014autonomous}, using deep reinforcement learning~\cite{jangir2020dynamic}, and designing a modifiable stochastic grammar~\cite{xiong2016robot}.

\begin{figure}[bh]
\begin{center}
    \includegraphics[width=0.95\linewidth]{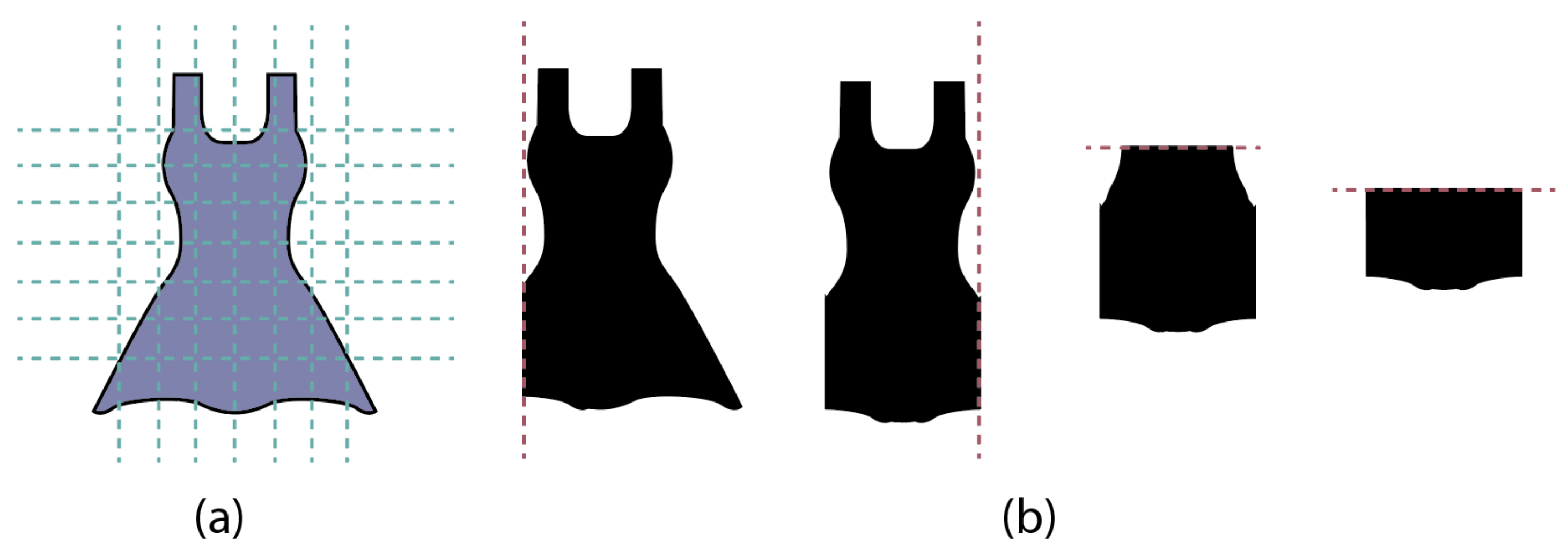}
\end{center}
  \caption{ (a) A dress with folding axes. (b) Folding steps.}
  \label{fig:fold}
\end{figure}

We abstract the clothes-folding challenge as a purely visual task: the contour of the dress/suit/shirt/pants is represented by a binary image, and folding clothes is characterised by manipulating images. Figure \ref{fig:fold} shows an image-like abstraction of folding a dress. 

The current state of the clothes $s$ is represented by a binary image $I$ from image space $\mathcal{S} =  \{0,1\}^{H\times W}$, and an action $a$ leads to fold the image along a certain axis (see figure \ref{fig:fold}). We also regard this task as a few-shot learning problem: as we are only given a few expert trajectories $\pi_{E} = \{\tau_{E_1}, \tau_{E_2},...,\tau_{E_{n_{e}}}\}$, where each trajectory $\tau_{E_i}$ is represented by the order sequence of states $(s_{E_{i1}}, s_{E_{i2}},...)$ towards the solution, the problem is how we can fold other arbitrary clothes we have not seen before.
\begin{figure}[th]
\begin{center}
    \includegraphics[width=0.85\linewidth]{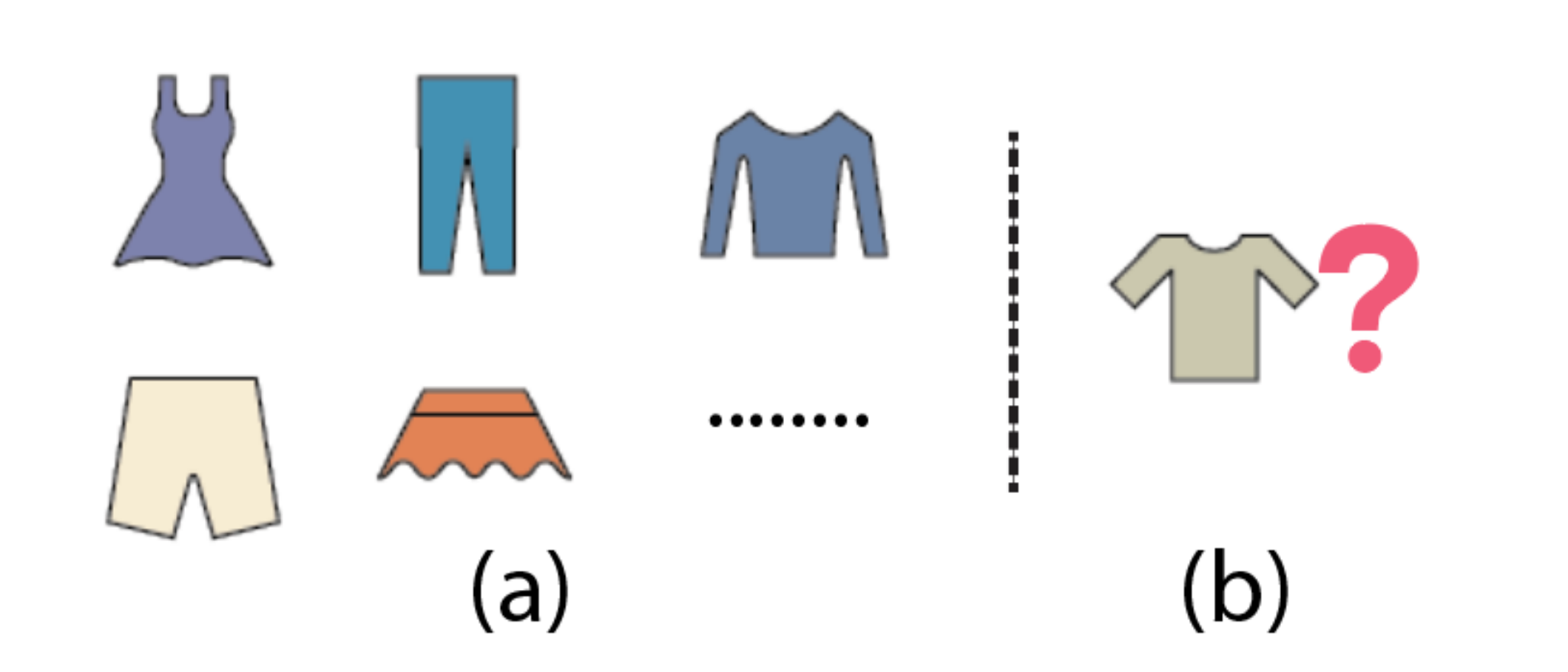}
\end{center}
  \caption{ (a) Expert sample clothes. (b) A T-shirt unseen before.}
\label{fig:foldq}
\end{figure}

We try several different ways to solve this task, including directly minimizing the CCL for expert trajectories and drawing on the popular algorithms from inverse reinforcement learning (IRL). The algorithms listed below can be applied not only to perform clothes-folding and furniture-organizing, but to solve a wide range of challenges related to robotics.

\begin{itemize}
    \item \textbf{Score learning} (SL): we can direclty give a score to a state $V_\delta: \mathcal{S}\mapsto [0,1]$, by learning from expert trajectories with the CCL (see equation \ref{eq:ccl}):
    \begin{align}\label{eq:vl}
        V_\delta(s) := f_\theta(s) .
    \end{align}
    \item \textbf{Max-entropy inverse reinforcement learning} (ME-IRL) \cite{ziebart2008maximum}: suppose a trajectory $\tau_{i}= (s_1,s_2,...)$ is sampled from the current cloth-folding policy $\pi_i$, and $F_{\psi}: \mathcal{S} \mapsto [0, 1]$, is the evaluation function for state $s$, we can calculate the gradient of $\psi$ by
    \begin{align}\label{eq:me-irl}
        \frac{\partial \mathcal{L}_\psi}{\partial \psi} = \mathbb{E}_{s \sim \tau_E}\bigg[\frac{\partial F_{\psi}(s)}{\partial \psi}\bigg] - \mathbb{E}_{s \sim \tau_i}\bigg[\frac{\partial F_{\psi}(s)}{\partial \psi}\bigg] , 
    \end{align}
    where $\mathcal{L}_\psi = P({\tau} | \pi_i, \tau \in \pi_{E})$ is the likelihood function of taking expert trajectories under the current policy.
    \item \textbf{Generative adversarial imitation learning} (GAIL)~\cite{ho2016generative}: after initializing the discriminator function $D_\omega: \mathcal{S} \mapsto [0,1]$ to distinguish states between expert and sampling trajectories, we can update $\omega$ with gradient
    \begin{align}\label{eq:gail}
        \frac{\partial \mathcal{L}_\omega}{\partial \omega} &= \mathbb{E}_{s \sim \tau_E}\bigg[\frac{\partial \log D_\omega(s)}{\partial \omega}\bigg] \nonumber \\
        &+ \mathbb{E}_{s \sim \tau_i}\bigg[\frac{\partial \log (1 - D_{\omega}(s))}{\partial \omega}\bigg]
    \end{align}
    where $\mathcal{L}_\omega$ is the adversarial loss \cite{ho2016generative} and $\tau_i$ shares the same meaning as above. Notice that we make a modification to GAIL by only distinguishing the state $s$ instead of the state-action pair $(s, a)$ since we are not given enough state-action pairs under few-shot settings.
\end{itemize}

\begin{figure}[th]
\begin{center}
    \includegraphics[width=0.85\linewidth]{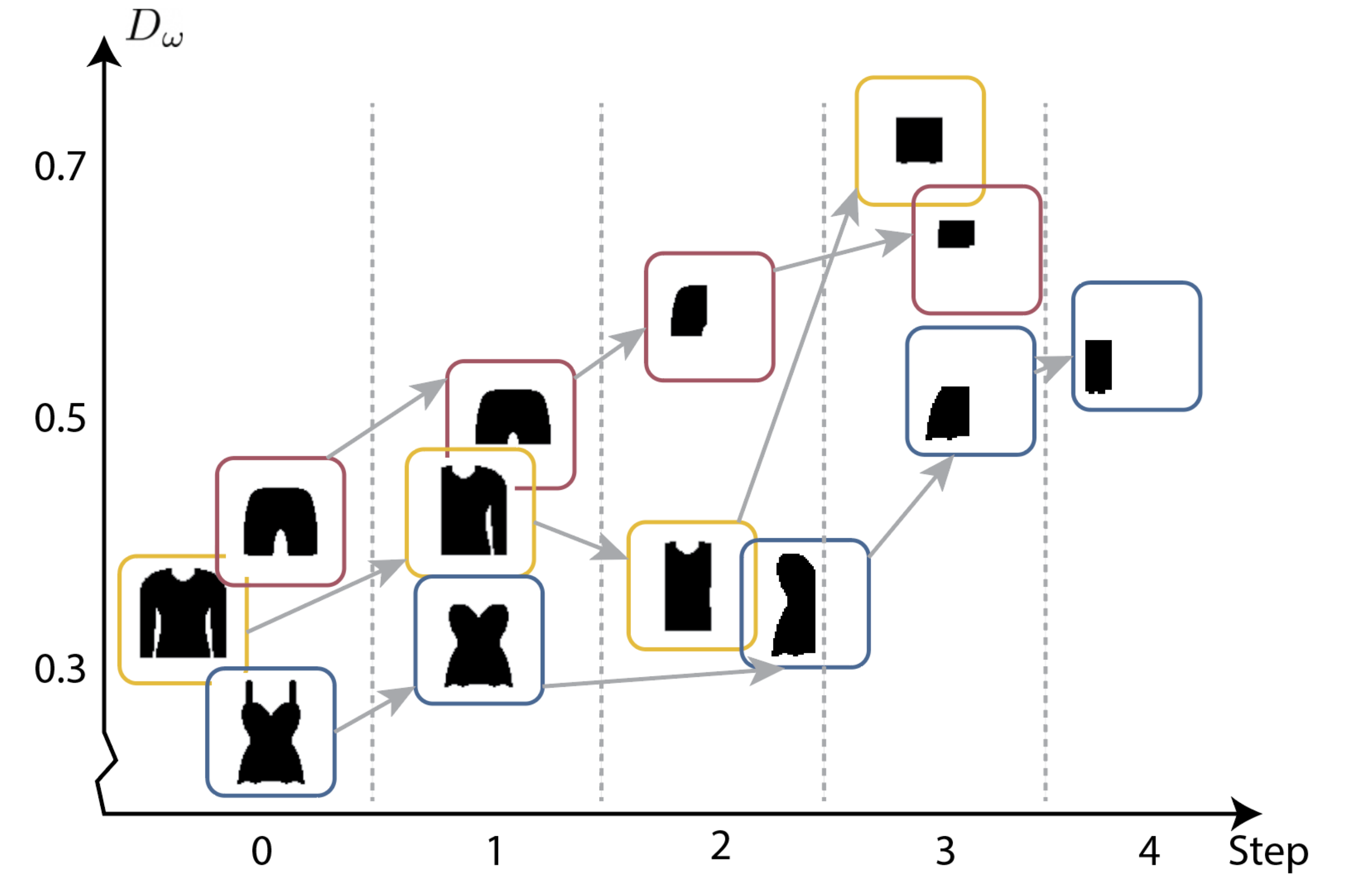}
\end{center}
  \caption{Aesthetic scores induced by $D_w$ (pre-trained).}
\label{figure:landscape}
\end{figure}

For simplicity, we regard the greedy policy deduced by the value of $V_\delta, F_\phi$ and $D_\omega$ as the propagated policy $\pi_i$ for SL, ME-IRL and GAIL. We assume that the clothes are put straight initially and they can only be folded along vertical and horizontal axes. The size of the image $I$ representing the state $s$ is $28\times 28$ and there are ten vertical and ten horizontal folding axes evenly distributed in the image.

We apply the network of the same structure in Section~\ref{sec:pre} \textit{Pre-training from the Tangram} for feature extraction to calculate $V_\delta, F_\psi$ and $D_\omega$. Three different ways along with pre-training or non-pre-training cases provide us with six different models. The models are trained on the expert trajectories from a total number of $18$ clothes, including dresses, long shirts, T-shirts, trousers, short pants, and skirts (three for each type). Then, models are tested on six new clothes from the aforementioned types and three clothes from other types. 

We refer to $V_\delta$,  $F_\phi$ and $D_\omega$ derived from equations \ref{eq:vl}, \ref{eq:me-irl}, and~\ref{eq:gail} as the \textit{aesthetic scores} of cloth-folding. Figure~\ref{figure:landscape} illustrates that $D_\omega$ increases as the clothes-folding process goes along. We compare the performance between different models by calculating the ranking of the ordered states $(s_{E_{i1}}, s_{E_{i2}},...)$ of expert trajectories based on $V_\delta$,  $F_\phi$ and $D_\omega$. Since on average the length of expert trajectories is around four, we only consider the precision at $K$ (P@$K$) with $K \leq 3$. Recall at $K$ as gives similar results.

Table \ref{table:tktrain} compares the overall difference in P@K between the pre-trained model and the non-pre-trained model (training from scratch) for the training expert samples (see the detailed comparison for each model in the Appendix). In general, we can see that pre-training improves the training precision and reduces the variance.  We select the best models of the six methods and test them once on the nice clothes that are unseen before. Table \ref{table:tktest} shows the mean and standard deviation of testing P@$K$. Except that ME-IRL without pre-training outperforms the pre-trained one w.r.t. P@$1$, pre-training improves the overall test accuracy, and the high precision on each value ($K = 1, 2, 3$) implicates overall better aesthetic scores.

ME-IRL and GAIL are common data-driven algorithms in the IRL domain. As with SL, their performance is heavily dependent on the amount of expert data given for training. Therefore, tuning from a pre-trained model can alleviate data reliance.

\begin{table}[thpb]
\centering
\begin{tabular}{c||ccc}
\hline 
                & \textbf{\small{P@1}} & \textbf{\small{P@2}}  & \textbf{\small{P@3}} \\ \hline \hline
\small\textbf{From scratch} & $0.54 \pm 0.5 $ & $0.66 \pm 0.32 $  & $0.76 \pm 0.21 $ \\
\small\textbf{Pre-training} & $0.77 \pm 0.41 $ & $0.84 \pm 0.26 $  & $0.86 \pm 0.18 $ \\ \hline
\end{tabular}
\newline
\caption{The mean and standard deviation of training P@$K$: a comparison between models with or without pre-training.}
\label{table:tktrain}
\end{table}
\begin{table}[tbh]
\centering
\begin{tabular}{c||c|c|c}
\hline        & \textbf{\small{P@1}} & \textbf{\small{P@2}} & \textbf{\small{P@3}} \\ \hline \hline
\small\textbf{SL}           &    $0.22 \small{\pm 0.46}$  &  $0.44 \pm 0.46$     & $0.55 \pm 0.47$\\
\small\textbf{SL(Pre)}       &   $\mathbf{0.89 \pm 0.33}$    &  $0.78 \pm 0.26$      &  $0.81 \pm 0.18$     \\ \hline
\small\textbf{ME-IRL}       &  $\mathbf{0.89 \pm 0.33}$     &  $0.78 \pm 0.26$     &  $0.74 \pm 0.22$     \\
\small\textbf{ME-IRL(Pre)} &   $0.67 \pm 0.50$    &  $\mathbf{0.94 \pm 0.17}$     &  $0.96 \pm 0.11$     \\ \hline
\small\textbf{GAIL}         &  $0.33 \pm 0.25$     &  $0.61 \pm 0.33$     &  $0.74 \pm 0.22$     \\
\small\textbf{GAIL(Pre)}     & $\mathbf{0.89 \pm 0.33}$     &  $\mathbf{0.94 \pm 0.17}$    &  $\mathbf{1.00 \pm 0.00}$     \\ \hline
\end{tabular}
\newline
\caption{The mean and standard deviation of testing P@$K$.}
\label{table:tktest}
\end{table}

\subsection{Evaluating room layouts} 
Generating room layouts is different from folding clothes in that the latter focuses on the shape change of a single object, while the former requires arranging multiple objects. These two pre-training exercises may correspond to the two variants of a replicating process of a tangram puzzle(see figure \ref{fig:loss}).

\begin{figure}[th]
\begin{center}
    \includegraphics[width=0.85\linewidth]{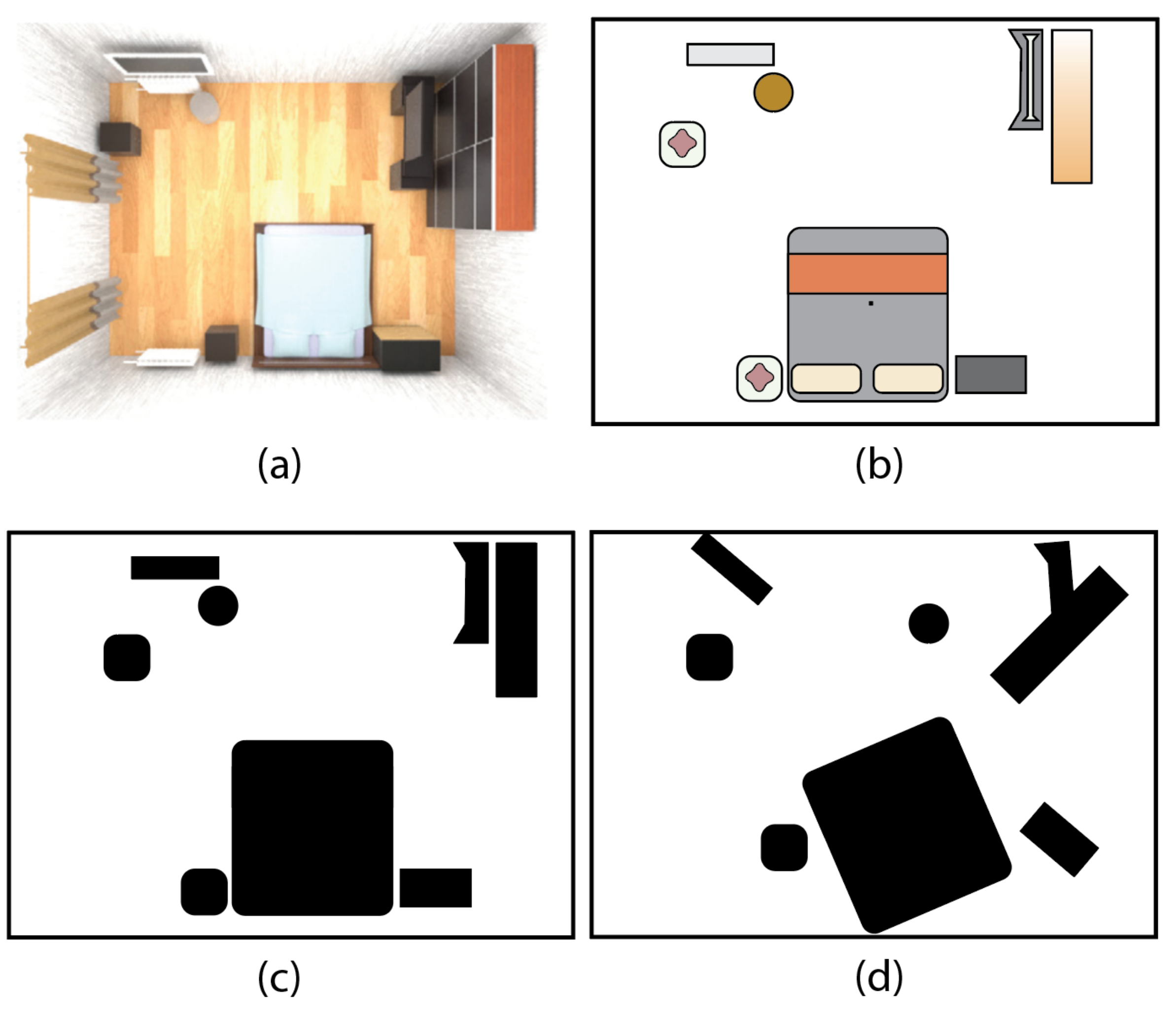}
\end{center}
  \caption{ (a) Original indoor scene sample from \cite{qi2018human}. (b) Abstract room layout. (c) Binary image representation. (d) Room messed up.}
\label{figure:furniture}
\end{figure}

The study of the layout generation has been active in various domains such as architectural design~\cite{nauata2020house, bao2013generating}  and game level design~\cite{ma2014game, hendrikx2013procedural}. We focus on the task of generating content for indoor scenes, especially furniture arrangement~\cite{yu2011make, ritchie2019fast, qi2018human}, and abstract it as a purely visual task as shown in Figure \ref{figure:furniture}.

We apply the state-of-the-art indoor scene synthesis using stochastic grammar~\cite{qi2018human} to generate the ground truth. Step by step, we perturb the room layout by the action $a$ that changes the position ($10$ pixels each step) and angle ($15^\circ$ each step) of the furniture, and the reversed steps generate an expert trajectory $\tau_{E_i}$ to tidy up the room.
\begin{table}[thpb]
\begin{tabular}{c||ccc}
\hline        & \textbf{\small{P@1}} & \textbf{\small{P@2}} & \textbf{\small{P@3}} \\ \hline \hline
\small\textbf{From scratch} & $0.18\pm 0.23$ & $0.23\pm 0.34$  & $0.32\pm 0.33$ \\
\small\textbf{Pre-training} & $0.23\pm 0.42$ & $0.28\pm 0.35$  & $0.39\pm 0.37$ \\ \hline
\end{tabular}
\newline
\caption{Training P @ $K$ comparison between models with or without pre-training.}
\label{table:roomtrain}
\end{table}

\begin{table}[hbtp]
\centering
\begin{tabular}{l||ll}
\hline
     & \textbf{\small{Original}} & \textbf{\small{Perturbed}} \\ \hline\hline
\textbf{\small{GAIL (from scratch)}} & $0.25 \pm 0.45$  & $0.23 \pm 0.38$     \\
\textbf{\small{GAIL (pre-trained)}} & $0.31 \pm 0.41$   & $0.29 \pm 0.35$    \\ \hline
\end{tabular}
\newline
\caption{Testing accuracy (P@$1$) of ranking the best room layout.}
\label{table:roomtest}
\end{table}
As in the previous experiment, we use a binary image $I$ to represent the current state $s$, and apply the three functions $V_\delta$, $F_\psi$ and $D_\omega$ to generate the aesthetic landscapes of the room. We only train our methods from $30$ generated expected trajectories and test them on $10$ groups of new room organizing trajectories.

Table \ref{table:roomtrain} shows the overall training improvement by pre-training. As in the previous experiment, pre-training improves the training accuracy. We select the best model GAIL from training, and we test it on identifying the best room layout from the testing trajectories. We also perturb each room in the trajectory a little to test the robustness of the model. Table \ref{table:roomtest} compares GAIL with/without pre-training on the testing challenges. The results indicate that pre-training on the Tangram improves performance in choosing the best room layout.
\subsection{Few-shot learning}
The goal of few-shot learning is to utilize new data having seen only a few samples. In this section, we focus on the $N$-way-$K$-shot classification: a typical problem to discriminate between $N$ classes with only $K$ samples from each to train from. 

The method we propose follows the paradigm of meta-learning \cite{sun2019meta}: we first train a feature extractor as a base-learner, which is later fine-tuned for another task through a meta-learner. As in previous experiments, a base learner is trained from the Tangram dataset, and then we perform a meta-test on the challenge of Omniglot~\cite{lake2019omniglot} and Multi-digit MNIST \cite{mulitdigitmnist}, where a binary image brings enough information to do classification. 

We select three methods: MAML~\cite{finn2017model}, ANIL~\cite{raghu2019rapid} and Prototypical Networks~\cite{snell2017prototypical} to train the meta-learner from our base-learner. MAML is a popular meta-learning algorithm for few-shot learning, achieving competitive performance on several benchmark few-shot learning problems. ANIL simplifies MAML by alleviating the inner training loop but keeping the training procedure for the task-specific part. Prototypical networks learn to map the prototypes to a metric space, and then distances between prototypes and encoded query inputs are used to make the classification.
To test the base-learner (feature extractor) trained on our Tangram data, we compare it with base-learners trained from EMNIST~\cite{cohen2017emnist} and Fashion-MNIST~\cite{xiao2017fashion}\footnote{we did not train the base-learner on MNIST\cite{deng2012mnist} because it is highly related to Multi-digit MNIST.}. All base-learners share the same network structure.

\begin{table}[hbtp] 
\centering
\begin{tabular}{l||cc}
\hline
              & \textbf{\small{Omniglot}} & \textbf{\small{Double-MNIST}} \\ \hline\hline
\textbf{\small{Random}}        & $33.7\% \pm 2.0\%$              & $7.3\% \pm 1.5\%$  \\ \hline
\textbf{\small{EMNIST}}        & $55.0\% \pm 5.4\%$                 & $26.8\% \pm 2.2\%$                    \\ \hline
\textbf{\small{Fashion-MNIST}} & $43.9\% \pm 4.1\%$    & $30.1\% \pm 1.2\%$  \\ \hline
\textbf{\small{Tangram}}       & $\mathbf{56.0\% \pm 4.7\%}$                & $\mathbf{36.0\% \pm 2.7\%}$                     \\ \hline
\end{tabular}
\newline
\caption{Five-way-five-shot learning: the mean and the standard deviation of testing accuracy (logistic regression only).}
\label{table:five-way-five-shot}
\end{table}

\begin{table}[hbtp]
\centering
\begin{tabular}{l|cc}
\hline
              & \textbf{\small{Omniglot}} & \textbf{\small{Double-MNIST}} \\ \hline\hline
\textbf{\small{Random}}    & $8.0\% \pm 0.7\%$              & $6.1\% \pm 0.1\%$  \\ \hline
\textbf{\small{EMNIST}}         & $\mathbf{22.1\% \pm 1.2\%}$                 & $7.5\% \pm 0.1\%$                    \\ \hline
\textbf{\small{Fashion-MNIST}} & $15.6\% \pm 1.4\%$    & $9.2\% \pm 0.5\%$                   \\ \hline
\textbf{\small{Tangram}}       & $22.0\% \pm 1.0\%$                & $\mathbf{10.5\% \pm 1.0\%}$                     \\ \hline
\end{tabular}
\newline
\caption{Twenty-way-five-shot learning: the mean and the standard deviation of testing accuracy (logistic regression only).}
\vspace{-2mm}
\label{table:twenty-way-five-shot}
\end{table}

\begin{figure*}[th]
\centering 
\includegraphics[width= 0.81\linewidth]{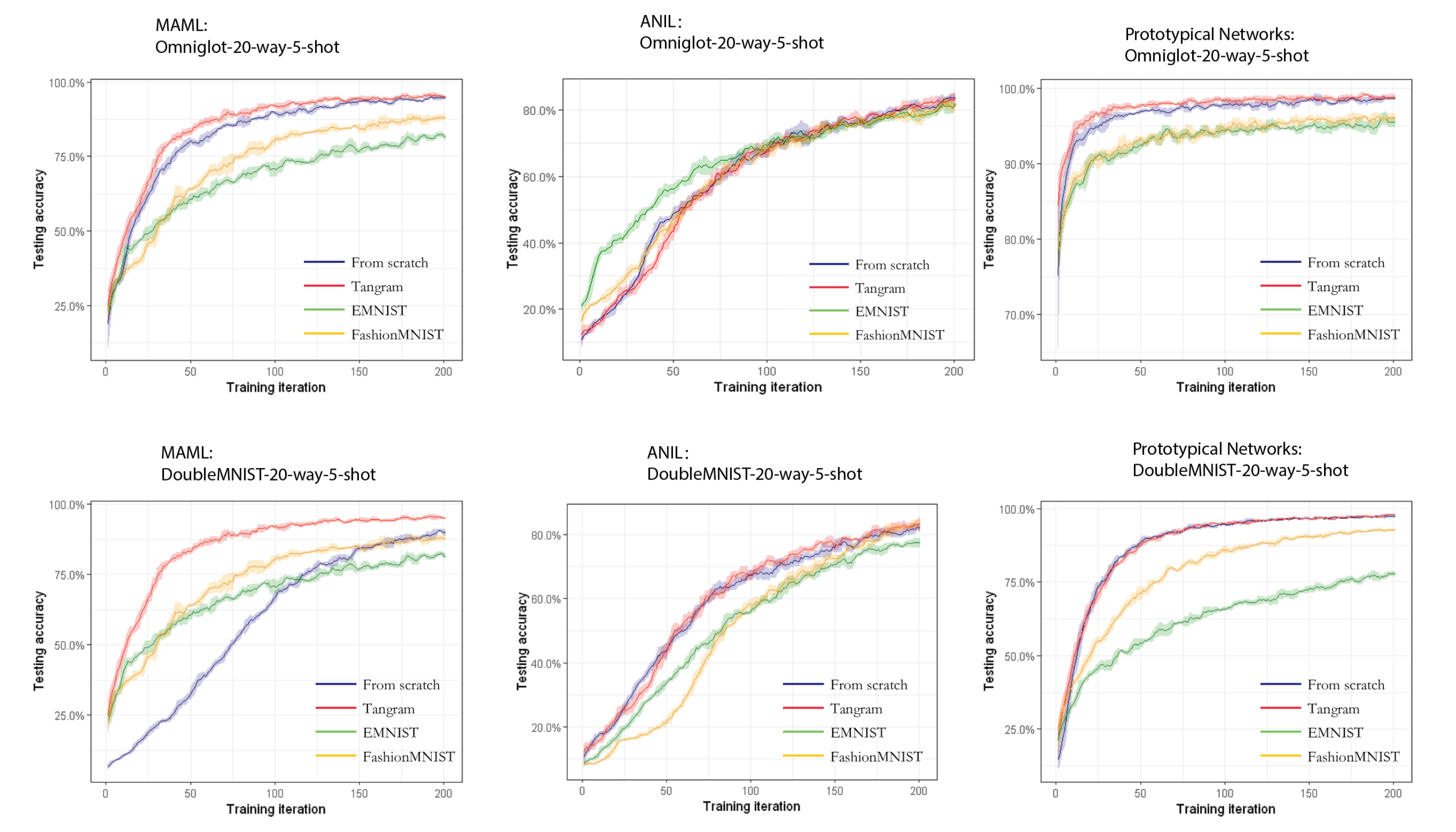}
\vspace{-2mm}
\caption{Testing accuracy of base-learners for different algorithms on different tasks.}
\vspace{-2mm}
\label{figure:tuning}
\end{figure*}

\begin{table*}[]
\centering
\begin{tabular}{l||cc|cc|cc}
\hline
    & \multicolumn{2}{c|}{\textbf{Flowers-17}}                        & \multicolumn{2}{c|}{\textbf{Flowers-102}}                       & \multicolumn{2}{c}{\textbf{Icons-50}}                           \\ \cline{2-7} 
    & \textit{\textbf{\small{ResNet-18}}} & \textit{\textbf{\small{EfficientNet-b0}}} & \textit{\textbf{\small{ResNet-18}}} & \textit{\textbf{\small{EfficientNet-b0}}} & \textit{\textbf{\small{ResNet-18}}} & \textit{\textbf{\small{EfficientNet-b0}}} \\ \hline \hline
\textbf{\small{From Scratch}}&    $73.5\% \pm 3.4\%$             & $76.1\% \pm 1.4\%$                        & $50.5\% \pm 1.3\%$                  & $51.7\% \pm 1.5\%$                        & $86.5\% \pm 0.4\%$                   & $84.5\% \pm 0.7\%$                         \\
\textbf{\small{Tangram}}      & $\mathbf{76.3\% \pm 3.8\%}$                  & $76.0\% \pm 1.2\%$                      & $\mathbf{51.1\% \pm 0.8\%}$                  & $50.6\% \pm 1.1\%$                        & $\mathbf{87.1\% \pm 1.1\%}$                  & $85.0\% \pm 1.0\%$                         \\ \hline
\end{tabular}
\newline
\caption{Classification results between training from scratch and pre-training from the Tangram. The inputs are binary images representing the contours only.}
\label{table:icontest}
\end{table*}

Before moving on to fine-tuning, we compare the feature extractors obtained by training on the above datasets. We train only the last layer of the network as logistic regression. As can be seen from Table \ref{table:five-way-five-shot} and Table \ref{table:twenty-way-five-shot}, feature extractors pre-trained on the Tangram, EMNIST, and Fashion-MNIST perform a lot better than the randomly initialized feature extractor. Except that the base-learner trained on EMNIST performs best in the 5-way-5-shot task on Omniglot, base-learners trained on the Tangram are powerful on other tasks, demonstrating their better adaptability.

Figure \ref{figure:tuning} compares the tuning process of different base-learners. Tuning the baser-learners pre-trained from the Tangram dataset guarantees the final performance compared with learning from scratch, while in some tasks it even speeds up convergence. However, for the other two feature extractors trained from EMNIST and FashionMNIST, although they may have a good start in some tasks, overall they tend to undermine the convergence speed and the final results, which reflects the difficulty of tuning a baser-learner for an irrelevant task. This result also demonstrates the importance of selecting a proper fundamental learning dataset in transfer learning. 

Table \ref{table:anil55} and Table \ref{table:anil205} compare the final training results between training from scratch and pre-training from Tangram, where we apply ANIL as the tuning algorithm. The results shown are trained after $500$ epochs. From the tables, we can see that pre-training from the Tangram provides slightly better results than training from scratch. 

\begin{table}[h]
\centering
\begin{tabular}{l||rr}
\hline
             & \multicolumn{1}{l}{\textbf{\small{Omiglot}}} & \multicolumn{1}{l}{\textbf{\small{Double MNIST}}} \\ \hline\hline
\textbf{\small{From scratch}} & $97.1\% \pm 1.4\%$ & $98.4\% \pm 1.3\%$  \\ \hline
\textbf{\small{Tangram}}      & $98.1\% \pm 1.0\%$ & $98.5\% \pm 1.0\%$  \\ \hline
\end{tabular}
\newline
\caption{Five-way-five-shot testing accuracy after training by ANIL.}
\label{table:anil55}
\end{table}

\begin{table}[hh]
\centering
\begin{tabular}{l||rr}
\hline
             & \multicolumn{1}{l}{\textbf{\small{Omiglot}}} & \multicolumn{1}{l}{\textbf{\small{Double MNIST}}} \\ \hline\hline
\textbf{\small{From scratch}} & $92.4\% \pm 1.0\%$ & $98.2\% \pm 0.3\%$  \\ \hline
\textbf{\small{Tangram}}      & $93.5\% \pm 0.9\%$ & $98.2\% \pm 0.2\%$  \\ \hline
\end{tabular}
\newline
\caption{Twenty-way-five-shot testing accuracy after training by ANIL.}
\vspace{-2mm}
\label{table:anil205}
\end{table}

\subsection{Icon recognition}

In this section, we study the recognition of abstract icons. While recognition tasks in natural pictures have been booming in the literature, visual abstraction receives comparably less attention. 

At first glance, icon recognition is a relatively straightforward task compared to the recognition task in natural images, since most icons are simple shapes that are not affected by light or blocking. However, it is worth considering how these abstract icons are formed, and how these seemingly simple icons can convey a variety of meanings. In this part, we wonder whether pre-training on the Tangram dataset assists in recognition of icons. Icons-50~\cite{hendrycks_2018} is a collection with $50$ types of icons, with training samples ranging from tens to hundreds for each type. We run the experiments with Icons-50 and test our methods on Flowers-17 and Flowers-102~\cite{nilsback2008automated}. 

\begin{figure}[th]
\begin{center}
    \includegraphics[width=0.80\linewidth]{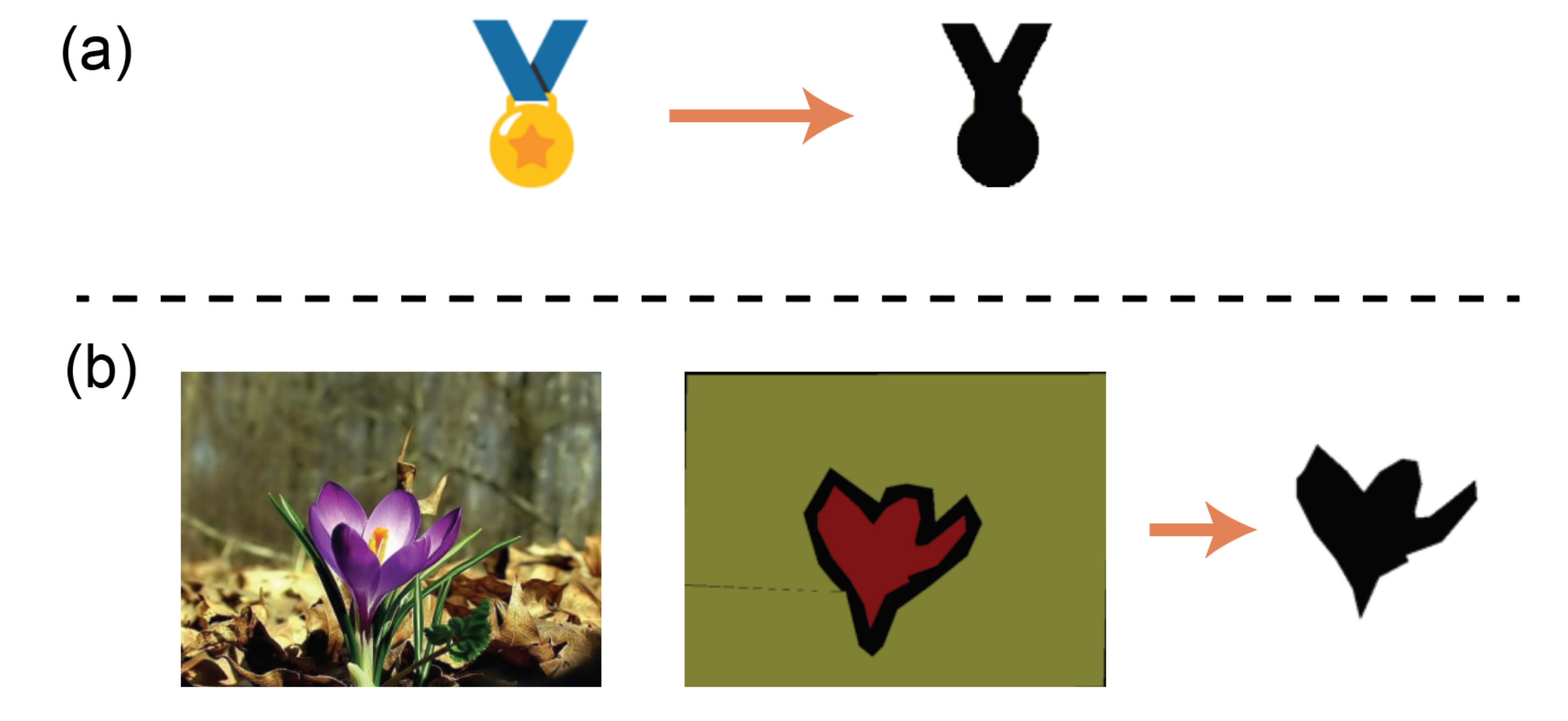}
\end{center}
  \caption{Data processing for (a) icons and (b) flowers.}
  \vspace{-1mm}
\label{fig:furniture}
\end{figure}

For Icons-50, we select icons with a white background coverage greater than $40\%$ and draw their contours, which results in a total number of $2,450$ samples. Flowers-17 and Flower-102 are well labeled with flower contours. Flowers-17 contains $17$ flower types and $849$ samples, and Flowers-102 has $102$ flower types and $8,189$ samples. For each dataset, $80\%$ of the samples are used for training and the remaining $20\%$ for testing. We use ResNet-18~\cite{he2016deep} and EfficientNet-b0~\cite{tan2019efficientnet} as the network architectures for icon classification. The inputs of the network are binary images of the size $224\times 224$. Table \ref{table:icontest} compares the model trained from scratch and the model pre-trained from Tangram. Although training Efficient-n0 from scratch brings good performance, the pre-trained model with ResNet-18 shows overall better testing accuracy. 
\section{Conclusion}

In this paper, we introduce the Tangram dataset that records step-by-step solutions to a tangram puzzle from human experience. The pre-training on the Tangram is applied to various tasks, including folding clothes, evaluation room layouts, few-shot learning challenges, and icon classification by contours. We hope that our pioneer work in abstract visual content could inspire the community to study visual anesthetics and image abstraction.

\bibliography{egbib.bib}

\begin{thebibliography}{55}
\providecommand{\natexlab}[1]{#1}

\bibitem[{Bao et~al.(2013)Bao, Yan, Mitra, and Wonka}]{bao2013generating}
Bao, F.; Yan, D.-M.; Mitra, N.~J.; and Wonka, P. 2013.
\newblock Generating and exploring good building layouts.
\newblock \emph{ACM Transactions on Graphics (TOG)}, 32(4): 1--10.

\bibitem[{Chakraborty, Gosthipaty, and Paul(2020)}]{chakraborty2020g}
Chakraborty, S.; Gosthipaty, A.~R.; and Paul, S. 2020.
\newblock G-SimCLR: Self-Supervised Contrastive Learning with Guided Projection
  via Pseudo Labelling.
\newblock \emph{arXiv preprint arXiv:2009.12007}.

\bibitem[{Chen et~al.(2020{\natexlab{a}})Chen, Radford, Child, Wu, Jun,
  Dhariwal, Luan, and Sutskever}]{chen2020generative}
Chen, M.; Radford, A.; Child, R.; Wu, J.; Jun, H.; Dhariwal, P.; Luan, D.; and
  Sutskever, I. 2020{\natexlab{a}}.
\newblock Generative pretraining from pixels.
\newblock In \emph{Proceedings of the 37th International Conference on Machine
  Learning}, volume~1.

\bibitem[{Chen et~al.(2020{\natexlab{b}})Chen, Kornblith, Norouzi, and
  Hinton}]{chen2020simple}
Chen, T.; Kornblith, S.; Norouzi, M.; and Hinton, G. 2020{\natexlab{b}}.
\newblock A simple framework for contrastive learning of visual
  representations.
\newblock \emph{arXiv preprint arXiv:2002.05709}.

\bibitem[{Cohen et~al.(2017)Cohen, Afshar, Tapson, and
  Van~Schaik}]{cohen2017emnist}
Cohen, G.; Afshar, S.; Tapson, J.; and Van~Schaik, A. 2017.
\newblock EMNIST: Extending MNIST to handwritten letters.
\newblock In \emph{2017 International Joint Conference on Neural Networks
  (IJCNN)}, 2921--2926. IEEE.

\bibitem[{Deng et~al.(2009)Deng, Dong, Socher, Li, Li, and
  Fei-Fei}]{imagenet_cvpr09}
Deng, J.; Dong, W.; Socher, R.; Li, L.-J.; Li, K.; and Fei-Fei, L. 2009.
\newblock {ImageNet: A Large-Scale Hierarchical Image Database}.
\newblock In \emph{CVPR09}.

\bibitem[{Deng(2012)}]{deng2012mnist}
Deng, L. 2012.
\newblock The mnist database of handwritten digit images for machine learning
  research [best of the web].
\newblock \emph{IEEE Signal Processing Magazine}, 29(6): 141--142.

\bibitem[{Deng, Loy, and Tang(2017)}]{deng2017image}
Deng, Y.; Loy, C.~C.; and Tang, X. 2017.
\newblock Image aesthetic assessment: An experimental survey.
\newblock \emph{IEEE Signal Processing Magazine}, 34(4): 80--106.

\bibitem[{Doshi, Shikkenawis, and Mitra(2019)}]{doshi2019image}
Doshi, N.; Shikkenawis, G.; and Mitra, S.~K. 2019.
\newblock Image Aesthetics Assessment Using Multi Channel Convolutional Neural
  Networks.
\newblock In \emph{International Conference on Computer Vision and Image
  Processing}, 15--24. Springer.

\bibitem[{Doumanoglou et~al.(2014)Doumanoglou, Kargakos, Kim, and
  Malassiotis}]{doumanoglou2014autonomous}
Doumanoglou, A.; Kargakos, A.; Kim, T.-K.; and Malassiotis, S. 2014.
\newblock Autonomous active recognition and unfolding of clothes using random
  decision forests and probabilistic planning.
\newblock In \emph{2014 IEEE international conference on robotics and
  automation (ICRA)}, 987--993. IEEE.

\bibitem[{Everingham et~al.(2010)Everingham, Van~Gool, Williams, Winn, and
  Zisserman}]{everingham2010pascal}
Everingham, M.; Van~Gool, L.; Williams, C.~K.; Winn, J.; and Zisserman, A.
  2010.
\newblock The pascal visual object classes (voc) challenge.
\newblock \emph{International journal of computer vision}, 88(2): 303--338.

\bibitem[{Felbo et~al.(2017)Felbo, Mislove, S{\o}gaard, Rahwan, and
  Lehmann}]{felbo2017using}
Felbo, B.; Mislove, A.; S{\o}gaard, A.; Rahwan, I.; and Lehmann, S. 2017.
\newblock Using millions of emoji occurrences to learn any-domain
  representations for detecting sentiment, emotion and sarcasm.
\newblock In \emph{Proceedings of the 2017 Conference on Empirical Methods in
  Natural Language Processing (EMNLP)}, 1615--1625.

\bibitem[{Finn, Abbeel, and Levine(2017)}]{finn2017model}
Finn, C.; Abbeel, P.; and Levine, S. 2017.
\newblock Model-Agnostic Meta-Learning for Fast Adaptation of Deep Networks.
\newblock In \emph{ICML}.

\bibitem[{Freeman(2007)}]{freeman2007complete}
Freeman, M. 2007.
\newblock \emph{The complete guide to light \& lighting in digital
  photography}.
\newblock Sterling Publishing Company, Inc.

\bibitem[{Haas(2014)}]{haas2014history}
Haas, J. 2014.
\newblock A history of the unity game engine.
\newblock \emph{Diss. WORCESTER POLYTECHNIC INSTITUTE}.

\bibitem[{He, Girshick, and Doll{\'a}r(2019)}]{he2019rethinking}
He, K.; Girshick, R.; and Doll{\'a}r, P. 2019.
\newblock Rethinking imagenet pre-training.
\newblock In \emph{Proceedings of the IEEE international conference on computer
  vision}, 4918--4927.

\bibitem[{He et~al.(2016)He, Zhang, Ren, and Sun}]{he2016deep}
He, K.; Zhang, X.; Ren, S.; and Sun, J. 2016.
\newblock Deep residual learning for image recognition.
\newblock In \emph{Proceedings of the IEEE conference on computer vision and
  pattern recognition}, 770--778.

\bibitem[{Hendrikx et~al.(2013)Hendrikx, Meijer, Van Der~Velden, and
  Iosup}]{hendrikx2013procedural}
Hendrikx, M.; Meijer, S.; Van Der~Velden, J.; and Iosup, A. 2013.
\newblock Procedural content generation for games: A survey.
\newblock \emph{ACM Transactions on Multimedia Computing, Communications, and
  Applications (TOMM)}, 9(1): 1--22.

\bibitem[{Hendrycks(2018)}]{hendrycks_2018}
Hendrycks, D. 2018.
\newblock Icons-50.

\bibitem[{Ho and Ermon(2016)}]{ho2016generative}
Ho, J.; and Ermon, S. 2016.
\newblock Generative adversarial imitation learning.
\newblock In \emph{Advances in neural information processing systems},
  4565--4573.

\bibitem[{Hu, Gripon, and Pateux(2020)}]{hu2020leveraging}
Hu, Y.; Gripon, V.; and Pateux, S. 2020.
\newblock Leveraging the Feature Distribution in Transfer-based Few-Shot
  Learning.
\newblock \emph{arXiv preprint arXiv:2006.03806}.

\bibitem[{Jangir, Aleny{\`a}, and Torras(2020)}]{jangir2020dynamic}
Jangir, R.; Aleny{\`a}, G.; and Torras, C. 2020.
\newblock Dynamic Cloth Manipulation with Deep Reinforcement Learning.
\newblock In \emph{2020 IEEE International Conference on Robotics and
  Automation (ICRA)}, 4630--4636. IEEE.

\bibitem[{Jing and Tian(2020)}]{jing2020self}
Jing, L.; and Tian, Y. 2020.
\newblock Self-supervised visual feature learning with deep neural networks: A
  survey.
\newblock \emph{IEEE Transactions on Pattern Analysis and Machine
  Intelligence}.

\bibitem[{Joshi et~al.(2011)Joshi, Datta, Fedorovskaya, Luong, Wang, Li, and
  Luo}]{joshi2011aesthetics}
Joshi, D.; Datta, R.; Fedorovskaya, E.; Luong, Q.-T.; Wang, J.~Z.; Li, J.; and
  Luo, J. 2011.
\newblock Aesthetics and emotions in images.
\newblock \emph{IEEE Signal Processing Magazine}, 28(5): 94--115.

\bibitem[{Karamatsu et~al.(2020)Karamatsu, Benitez-Garcia, Yanai, and
  Uchida}]{karamatsu2020iconify}
Karamatsu, T.; Benitez-Garcia, G.; Yanai, K.; and Uchida, S. 2020.
\newblock Iconify: Converting Photographs into Icons.
\newblock In \emph{Proceedings of the 2020 Joint Workshop on Multimedia
  Artworks Analysis and Attractiveness Computing in Multimedia}, 7--12.

\bibitem[{Ke, Tang, and Jing(2006)}]{ke2006design}
Ke, Y.; Tang, X.; and Jing, F. 2006.
\newblock The design of high-level features for photo quality assessment.
\newblock In \emph{2006 IEEE Computer Society Conference on Computer Vision and
  Pattern Recognition (CVPR'06)}, volume~1, 419--426. IEEE.

\bibitem[{Lagunas, Garces, and Gutierrez(2019)}]{lagunas2019learning}
Lagunas, M.; Garces, E.; and Gutierrez, D. 2019.
\newblock Learning icons appearance similarity.
\newblock \emph{Multimedia Tools and Applications}, 78(8): 10733--10751.

\bibitem[{Lake, Salakhutdinov, and Tenenbaum(2019)}]{lake2019omniglot}
Lake, B.~M.; Salakhutdinov, R.; and Tenenbaum, J.~B. 2019.
\newblock The Omniglot challenge: a 3-year progress report.
\newblock \emph{Current Opinion in Behavioral Sciences}, 29: 97--104.

\bibitem[{Land and Nilsson(2012)}]{land2012animal}
Land, M.~F.; and Nilsson, D.-E. 2012.
\newblock \emph{Animal eyes}.
\newblock Oxford University Press.

\bibitem[{Lin et~al.(2014)Lin, Maire, Belongie, Hays, Perona, Ramanan,
  Doll{\'a}r, and Zitnick}]{lin2014microsoft}
Lin, T.-Y.; Maire, M.; Belongie, S.; Hays, J.; Perona, P.; Ramanan, D.;
  Doll{\'a}r, P.; and Zitnick, C.~L. 2014.
\newblock Microsoft coco: Common objects in context.
\newblock In \emph{European conference on computer vision}, 740--755. Springer.

\bibitem[{Lu et~al.(2015)Lu, Lin, Jin, Yang, and Wang}]{lu2015rating}
Lu, X.; Lin, Z.; Jin, H.; Yang, J.; and Wang, J.~Z. 2015.
\newblock Rating image aesthetics using deep learning.
\newblock \emph{IEEE Transactions on Multimedia}, 17(11): 2021--2034.

\bibitem[{Luo, Wang, and Tang(2011)}]{luo2011content}
Luo, W.; Wang, X.; and Tang, X. 2011.
\newblock Content-based photo quality assessment.
\newblock In \emph{2011 International Conference on Computer Vision},
  2206--2213. IEEE.

\bibitem[{Ma et~al.(2014)Ma, Vining, Lefebvre, and Sheffer}]{ma2014game}
Ma, C.; Vining, N.; Lefebvre, S.; and Sheffer, A. 2014.
\newblock Game level layout from design specification.
\newblock In \emph{Computer Graphics Forum}, volume~33, 95--104. Wiley Online
  Library.

\bibitem[{Madan et~al.(2018)Madan, Bylinskii, Tancik, Recasens, Zhong,
  Alsheikh, Pfister, Oliva, and Durand}]{madan2018synthetically}
Madan, S.; Bylinskii, Z.; Tancik, M.; Recasens, A.; Zhong, K.; Alsheikh, S.;
  Pfister, H.; Oliva, A.; and Durand, F. 2018.
\newblock Synthetically trained icon proposals for parsing and summarizing
  infographics.
\newblock \emph{arXiv preprint arXiv:1807.10441}.

\bibitem[{Nauata et~al.(2020)Nauata, Chang, Cheng, Mori, and
  Furukawa}]{nauata2020house}
Nauata, N.; Chang, K.-H.; Cheng, C.-Y.; Mori, G.; and Furukawa, Y. 2020.
\newblock House-GAN: Relational Generative Adversarial Networks for
  Graph-constrained House Layout Generation.
\newblock \emph{arXiv preprint arXiv:2003.06988}.

\bibitem[{Ni et~al.(2017)Ni, Ma, Zeng, Chen, Cai, and Ma}]{ni2017esim}
Ni, Z.; Ma, L.; Zeng, H.; Chen, J.; Cai, C.; and Ma, K.-K. 2017.
\newblock ESIM: Edge similarity for screen content image quality assessment.
\newblock \emph{IEEE Transactions on Image Processing}, 26(10): 4818--4831.

\bibitem[{Nilsback and Zisserman(2008)}]{nilsback2008automated}
Nilsback, M.-E.; and Zisserman, A. 2008.
\newblock Automated flower classification over a large number of classes.
\newblock In \emph{2008 Sixth Indian Conference on Computer Vision, Graphics \&
  Image Processing}, 722--729. IEEE.

\bibitem[{Nishiyama et~al.(2011)Nishiyama, Okabe, Sato, and
  Sato}]{nishiyama2011aesthetic}
Nishiyama, M.; Okabe, T.; Sato, I.; and Sato, Y. 2011.
\newblock Aesthetic quality classification of photographs based on color
  harmony.
\newblock In \emph{CVPR 2011}, 33--40. IEEE.

\bibitem[{Orsic et~al.(2019)Orsic, Kreso, Bevandic, and
  Segvic}]{orsic2019defense}
Orsic, M.; Kreso, I.; Bevandic, P.; and Segvic, S. 2019.
\newblock In defense of pre-trained imagenet architectures for real-time
  semantic segmentation of road-driving images.
\newblock In \emph{Proceedings of the IEEE conference on computer vision and
  pattern recognition}, 12607--12616.

\bibitem[{Pennington, Socher, and Manning(2014)}]{pennington2014glove}
Pennington, J.; Socher, R.; and Manning, C.~D. 2014.
\newblock Glove: Global vectors for word representation.
\newblock In \emph{Proceedings of the 2014 conference on empirical methods in
  natural language processing (EMNLP)}, 1532--1543.

\bibitem[{Qi et~al.(2018)Qi, Zhu, Huang, Jiang, and Zhu}]{qi2018human}
Qi, S.; Zhu, Y.; Huang, S.; Jiang, C.; and Zhu, S.-C. 2018.
\newblock Human-centric indoor scene synthesis using stochastic grammar.
\newblock In \emph{Proceedings of the IEEE Conference on Computer Vision and
  Pattern Recognition}, 5899--5908.

\bibitem[{Raghu et~al.(2019)Raghu, Raghu, Bengio, and Vinyals}]{raghu2019rapid}
Raghu, A.; Raghu, M.; Bengio, S.; and Vinyals, O. 2019.
\newblock Rapid Learning or Feature Reuse? Towards Understanding the
  Effectiveness of MAML.
\newblock In \emph{International Conference on Learning Representations}.

\bibitem[{Ritchie, Wang, and Lin(2019)}]{ritchie2019fast}
Ritchie, D.; Wang, K.; and Lin, Y.-a. 2019.
\newblock Fast and flexible indoor scene synthesis via deep convolutional
  generative models.
\newblock In \emph{Proceedings of the IEEE Conference on Computer Vision and
  Pattern Recognition}, 6182--6190.

\bibitem[{Shinya, Simo-Serra, and Suzuki(2019)}]{shinya2019understanding}
Shinya, Y.; Simo-Serra, E.; and Suzuki, T. 2019.
\newblock Understanding the Effects of Pre-Training for Object Detectors via
  Eigenspectrum.
\newblock In \emph{Proceedings of the IEEE International Conference on Computer
  Vision Workshops}, 0--0.

\bibitem[{Snell, Swersky, and Zemel(2017)}]{snell2017prototypical}
Snell, J.; Swersky, K.; and Zemel, R. 2017.
\newblock Prototypical networks for few-shot learning.
\newblock In \emph{Advances in neural information processing systems},
  4077--4087.

\bibitem[{Sun et~al.(2019{\natexlab{a}})Sun, Liu, Chua, and
  Schiele}]{sun2019meta}
Sun, Q.; Liu, Y.; Chua, T.-S.; and Schiele, B. 2019{\natexlab{a}}.
\newblock Meta-transfer learning for few-shot learning.
\newblock In \emph{Proceedings of the IEEE conference on computer vision and
  pattern recognition}, 403--412.

\bibitem[{Sun(2019)}]{mulitdigitmnist}
Sun, S.-H. 2019.
\newblock Multi-digit MNIST for Few-shot Learning.

\bibitem[{Sun et~al.(2019{\natexlab{b}})Sun, Min, Zhai, Gu, Duan, and
  Ma}]{sun2019mc360iqa}
Sun, W.; Min, X.; Zhai, G.; Gu, K.; Duan, H.; and Ma, S. 2019{\natexlab{b}}.
\newblock MC360IQA: A Multi-channel CNN for Blind 360-Degree Image Quality
  Assessment.
\newblock \emph{IEEE Journal of Selected Topics in Signal Processing}, 14(1):
  64--77.

\bibitem[{Tan and Le(2019)}]{tan2019efficientnet}
Tan, M.; and Le, Q.~V. 2019.
\newblock Efficientnet: Rethinking model scaling for convolutional neural
  networks.
\newblock \emph{arXiv preprint arXiv:1905.11946}.

\bibitem[{Xiao, Rasul, and Vollgraf(2017)}]{xiao2017fashion}
Xiao, H.; Rasul, K.; and Vollgraf, R. 2017.
\newblock Fashion-mnist: a novel image dataset for benchmarking machine
  learning algorithms.
\newblock \emph{arXiv preprint arXiv:1708.07747}.

\bibitem[{Xiong et~al.(2016)Xiong, Shukla, Xiong, and Zhu}]{xiong2016robot}
Xiong, C.; Shukla, N.; Xiong, W.; and Zhu, S.-C. 2016.
\newblock Robot learning with a spatial, temporal, and causal and-or graph.
\newblock In \emph{2016 IEEE International Conference on Robotics and
  Automation (ICRA)}, 2144--2151. IEEE.

\bibitem[{Yang et~al.(2015)Yang, Li, Fermuller, and Aloimonos}]{yang2015robot}
Yang, Y.; Li, Y.; Fermuller, C.; and Aloimonos, Y. 2015.
\newblock Robot learning manipulation action plans by" watching" unconstrained
  videos from the world wide web.
\newblock In \emph{Twenty-ninth AAAI conference on artificial intelligence}.
  Citeseer.

\bibitem[{Yu et~al.(2011)Yu, Yeung, Tang, Terzopoulos, Chan, and
  Osher}]{yu2011make}
Yu, L.~F.; Yeung, S.~K.; Tang, C.~K.; Terzopoulos, D.; Chan, T.~F.; and Osher,
  S.~J. 2011.
\newblock Make it home: automatic optimization of furniture arrangement.
\newblock \emph{ACM Transactions on Graphics (TOG)-Proceedings of ACM SIGGRAPH
  2011, v. 30,(4), July 2011, article no. 86}, 30(4).

\bibitem[{Zhai and Min(2020)}]{zhai2020perceptual}
Zhai, G.; and Min, X. 2020.
\newblock Perceptual image quality assessment: a survey.
\newblock \emph{SCIENCE CHINA Information Sciences}, 63(11): 211301.

\bibitem[{Ziebart et~al.(2008)Ziebart, Maas, Bagnell, and
  Dey}]{ziebart2008maximum}
Ziebart, B.~D.; Maas, A.~L.; Bagnell, J.~A.; and Dey, A.~K. 2008.
\newblock Maximum entropy inverse reinforcement learning.
\newblock In \emph{Aaai}, volume~8, 1433--1438. Chicago, IL, USA.

\end{thebibliography}

\newpage
\section*{Appendix A: Pre-training Network}
Networks for pre-training, folding clothes, organizing furniture and Omniglot/Multi-digit MNIST challenge share the same structure as feature extractors. The following figure plots the detailed structure of the feature extractor.  The feature extractor has about $112$k trainable parameters.
\begin{figure}[h]
    \centering
    \includegraphics[width=0.8\linewidth]{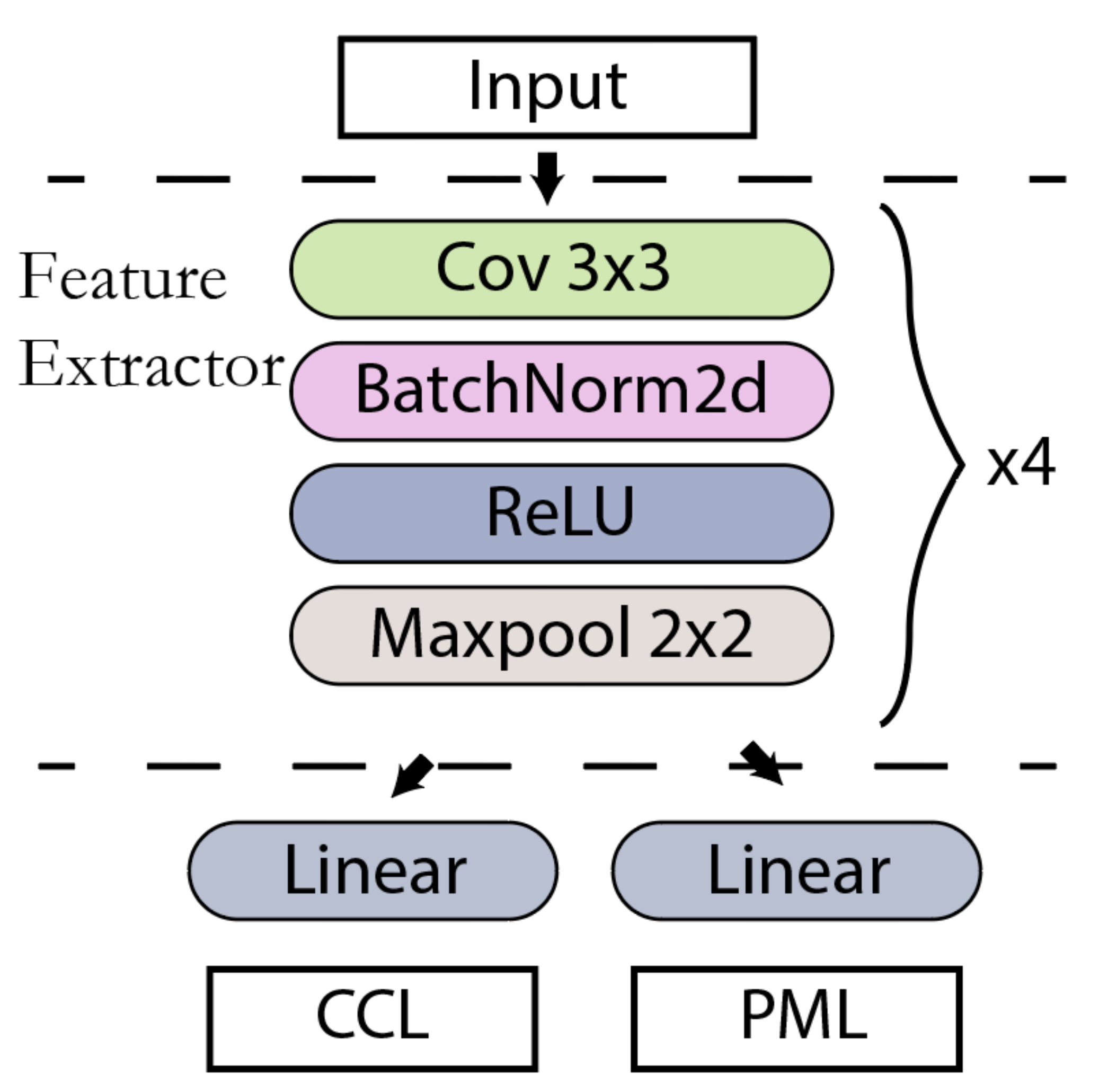}
    \caption{Network architecture for pre-training}
    \label{fig:my_label}
\end{figure}

\section*{Appendix B: The Tangram Collection}
The puzzles collected are mainly from \href{https://www.tangram-channel.com/}{Tangram Channel}. We have also published the labeling tool online by \href{https://github.com/}{Github} and \href{https://get.webgl.org/}{WebGL}. To help us label more data and contribute to research, please visit \url{https://yizhouzhao.github.io/TangramLabeling}. 

To understand how we label from the \href{https://github.com/}{Github} host, please refer to the source code of the labeling tool \url{https://github.com/yizhouzhao/TangramLabeling}.

\begin{figure}[h]
    \centering
    \includegraphics[width=0.95\linewidth]{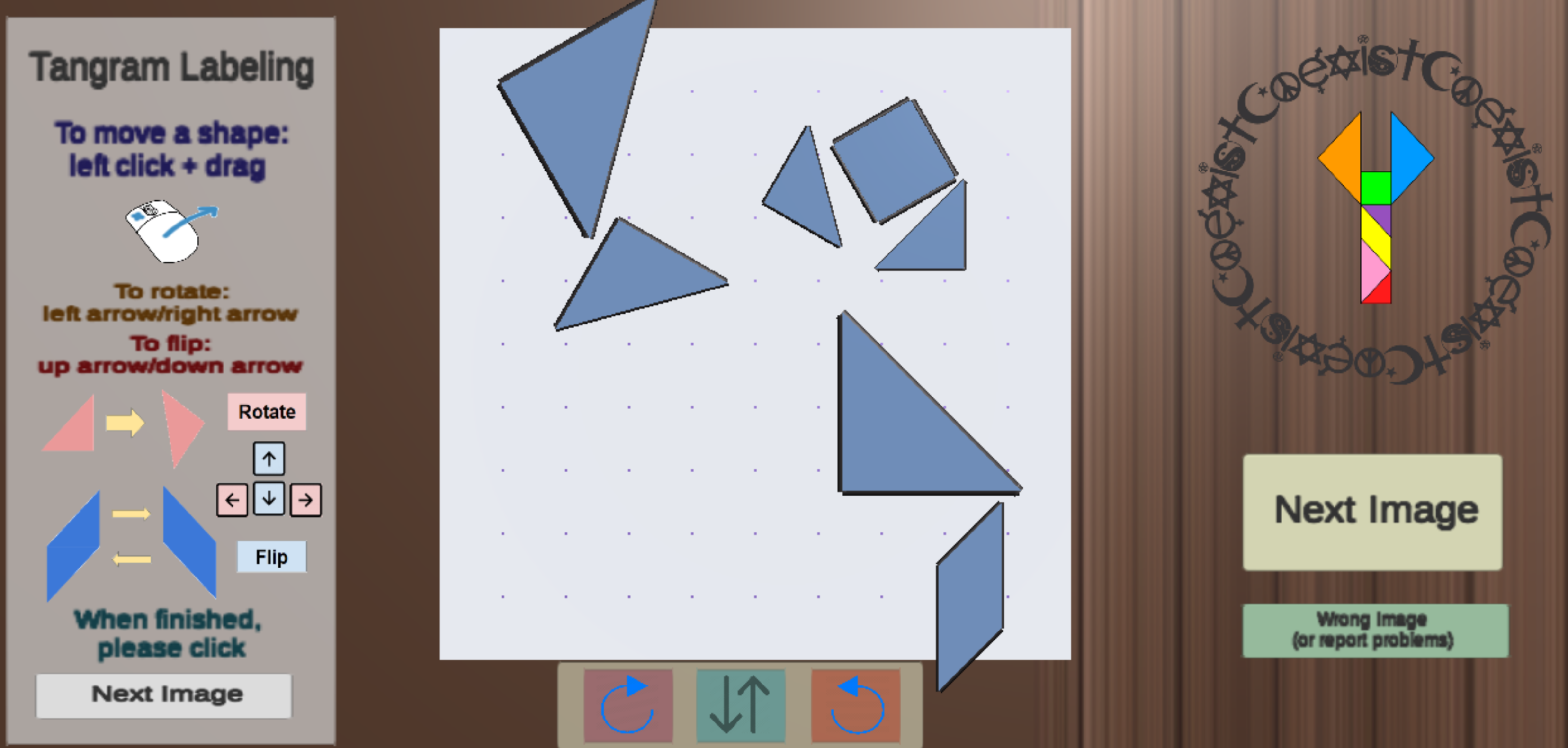}
    \caption{Tangram labeling tool.}
    \label{fig:my_label}
\end{figure}

The above image shows the tool for labeling tangram puzzles, where a board in the middle contains $100\times 100$ grid points and seven tans. One tan can be moved on the grid point, be rotated every $15^\circ$ and be flipped vertically. Every step of the process is recorded. 

\section*{Appendix C: Clothes-Folding details}
We use a total number of $18$ clothes with their folding steps for training, including trousers, dresses, T-shirts, skirts, pants and jackets. The clothes were all drawn by hand on \href{https://www.adobe.com/}{Adobe Illustrator}.
\begin{figure}[h]
    \centering
    \includegraphics[width=0.95\linewidth]{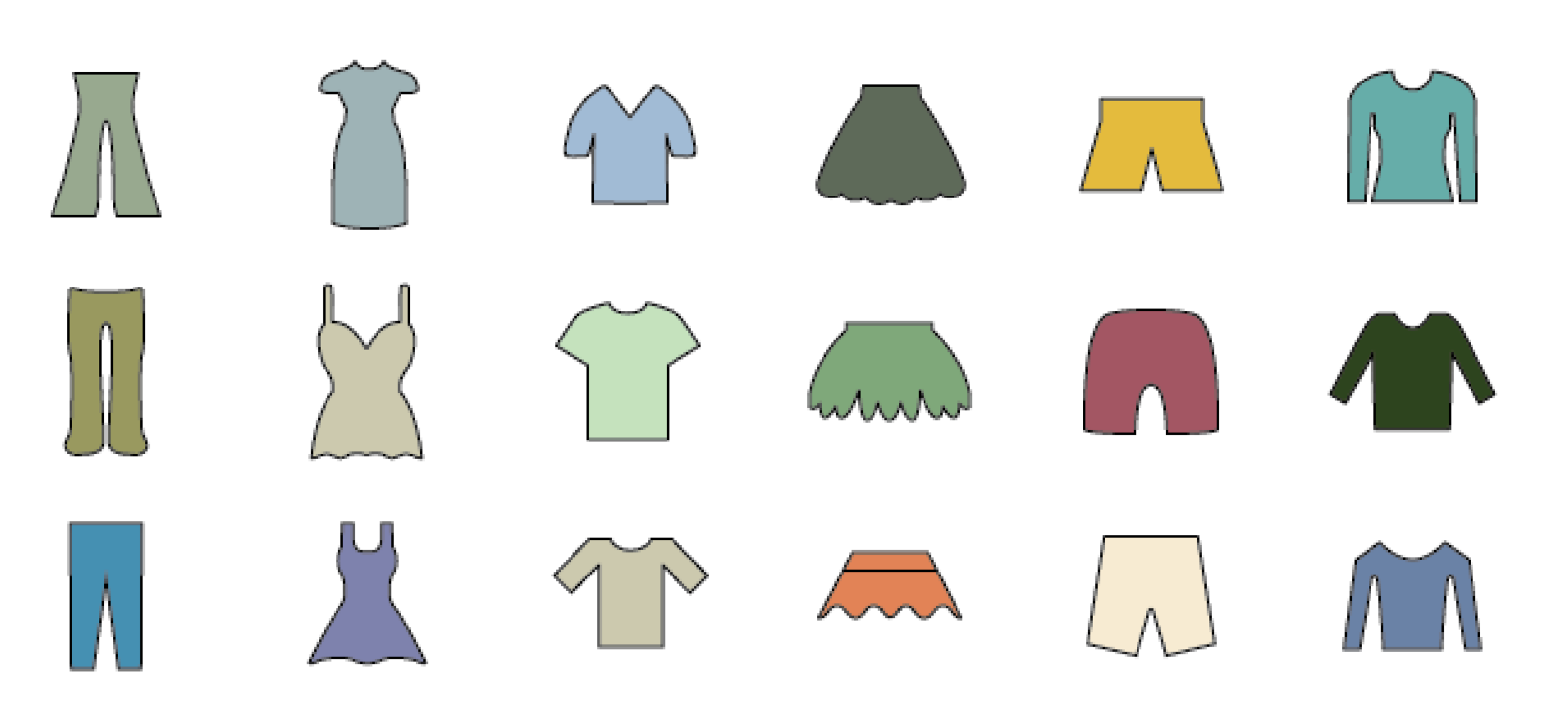}
    \caption{Training clothes}
    \label{fig:my_label}
\end{figure}

The following figure plots nine clothes used for testing, including three vests which do not belong to any type of training clothes.

\begin{figure}[h]
    \centering
    \includegraphics[width=0.95\linewidth]{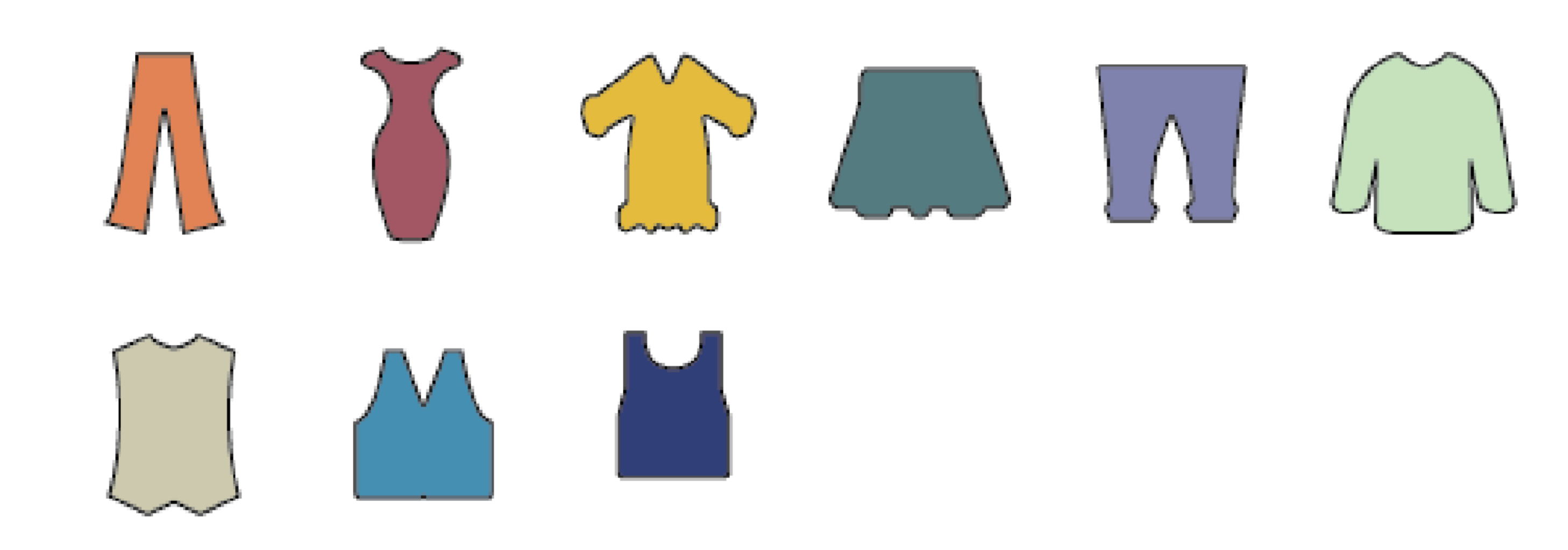}
    \caption{Testing clothes}
    \label{fig:my_label}
\end{figure}

The reason that we call $V_\delta$, $F_\phi$ and $D_\omega$ aesthetic landscapes instead of value functions is they give a intuitive guidance for aesthetic tasks. A greedy policy based on them can generate results good enough. Moreover, because we have not clearly defined the rewards and they are not calculated by Bellman equation, to call them value functions seem inappropriate. An example (folding a pair of short pants) with the aesthetic landscape is illustrated in the following table.

Generally, GAIL with pre-training on our Tangram generates reasonable landscapes.

\begin{table}[h]
\begin{tabular}{c||cccc}
\hline
\textbf{}            & \textbf{Step 0} & \textbf{Step 1} & \textbf{Step 2} & \textbf{Step 3} \\ \hline\hline
\textbf{SL}          & 0.4643          & 0.4673          & 0.5434          & 0.4979          \\
\textbf{SL(Pre)}     & 0.5015          & 0.4608          & 0.4967          & 0.5063          \\ \hline
\textbf{ME-IRL}      & 0.0001        & 0.0004          & 0.9722          & 0.9999          \\
\textbf{ME-IRL(Pre)} & 0.0014          & 0.0081          & 0.0183          & 0.033           \\ \hline
\textbf{GAIL}        & 0.4783          & 0.4039          & 0.4886          & 0.5071          \\
\textbf{GAIL(Pre)}   & 0.4117          & 0.4938          & 0.6186          & 0.6727          \\ \hline
\end{tabular}
\newline
\caption{Aesthetic landscapes when folding a pair of short pants.}
\end{table}

\section*{Appendix D: Organizing room layouts}
We generated $40$ room layouts including living room, bedrooms, kitchens and bathrooms from \textit{Human-centric Indoor Scene Synthesis} \cite{qi2018human}, by ruling out the following configurations: \textit{door}, \textit{rug}, \textit{ottoman}, \textit{cutting board}, \textit{fence}, \textit{clock}, \textit{vase}, \textit{television}, \textit{partition}, \textit{person}, \textit{garage door}, \textit{picture frame}, \textit{toy}, and \textit{shelving}. The following figure plots some samples.

We use $30$ of them for training and $10$ for testing.

We generate expert trajectories by randomly moving and rotating the furniture in the room. In each step, one desk/chair/TV stand/... can be moved $10$ pixels randomly to left/right/up/down, or be rotated by $15^\circ$ clockwise/counter-clockwise from its center. On average, there are $15$ types of furniture in one room therefore the average length of expert trajectories is $15$. We apply the same method as cloth-folding to solve aesthetic evaluation for room layouts.

\begin{figure}[h]
    \centering
    \includegraphics[width=0.95\linewidth]{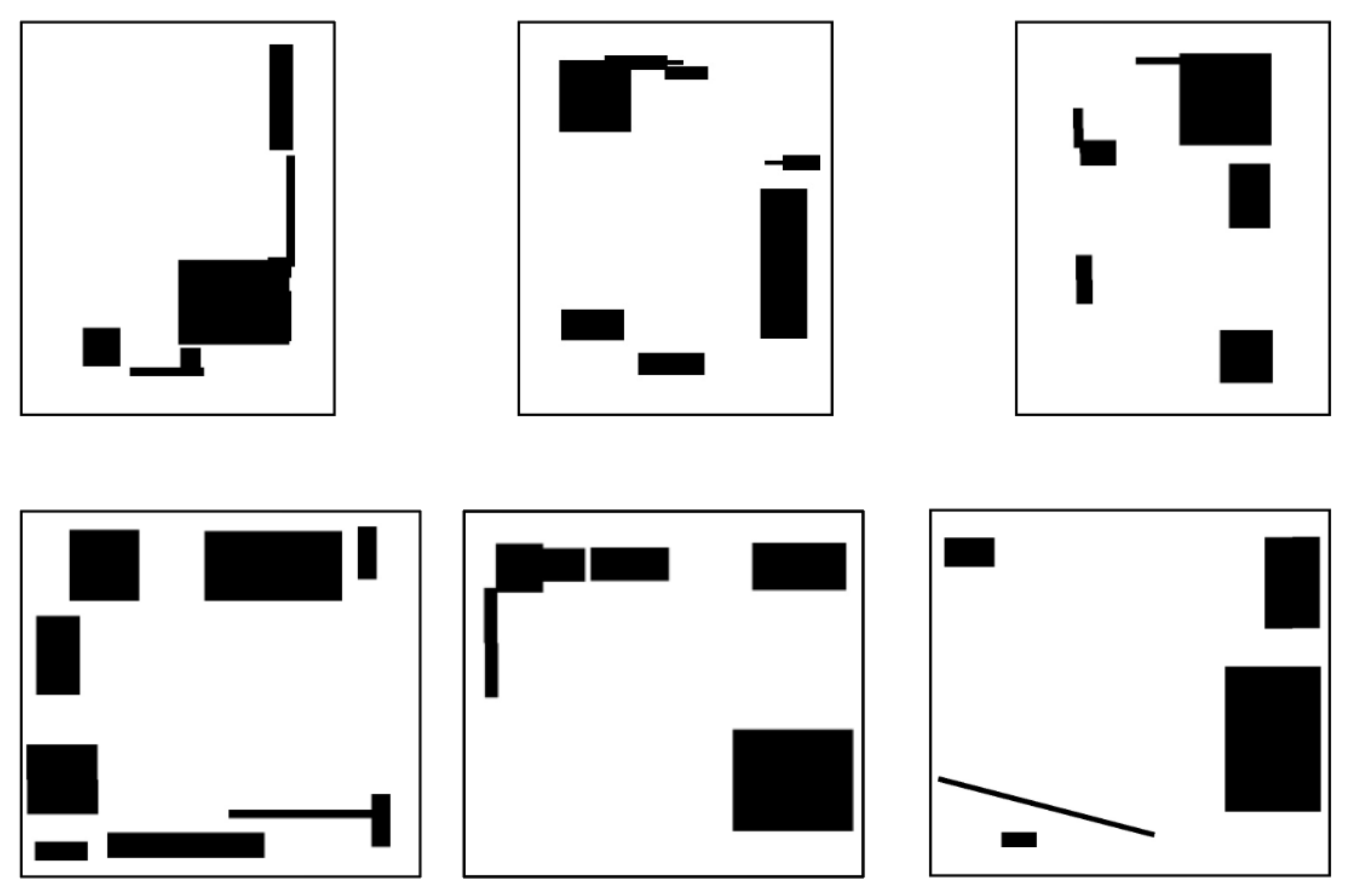}
    \caption{Room layout samples}
    \label{fig:my_label}
\end{figure}

The following tables show the detailed mean and standard deviation of the precision in the training. Except that ME-IRL performs better if it is trained from scratch w.r.t. P @ $1$, methods incorporating with pre-training on our Tangram generally perform better.

\begin{table}[h!]
\centering
\begin{tabular}{c||ccc}
\hline
                 & \textbf{P @ 1} & \textbf{P @ 2} & \textbf{P @ 3} \\ \hline\hline
\textbf{SL-Mean} & 0.54           & 0.659          & 0.767          \\
\textbf{SL-SD}   & 0.252          & 0.103          & 0.045          \\ \hline
\end{tabular}
\newline
\caption{Mean and standard deviation of Score learning.}
\end{table}

\begin{table}[h!]
\centering
\begin{tabular}{c||ccc}
\hline
 & \textbf{P @ 1} &{\textbf{P @ 2}} & {\textbf{P @ 3}} \\ \hline\hline
\textbf{SL(Pre)-Mean} & 0.778                              & 0.849                              & 0.862                              \\
\textbf{SL(Pre)-SD}   & 0.176                              & 0.07                               & 0.035                              \\ \hline
\end{tabular}
\newline
\caption{Mean and standard deviation of Score learning (Pre-trained).}
\end{table}

\begin{table}[h!]
\centering
\begin{tabular}{c||ccc}
\hline
                 & \textbf{P @ 1} & \textbf{P @ 2} & \textbf{P @ 3} \\ \hline\hline
\textbf{ME-Mean} & 1              & 0.905          & 0.841          \\
\textbf{ME-SD}   & 0              & 0.201          & 0.201          \\ \hline
\end{tabular}
\newline
\caption{Mean and standard deviation of Max-entropy IRL.}
\end{table}

\begin{table}[h!]
\centering
\begin{tabular}{c||ccc}
\hline
                      & \textbf{P @ 1} & \textbf{P @ 2} & \textbf{P @ 3} \\ \hline\hline
\textbf{ME(Pre)-Mean} & 0.952          & 0.976          & 0.937          \\
\textbf{ME(Pre)-SD}   & 0.218          & 0.109          & 0.134          \\ \hline
\end{tabular}
\newline
\caption{Mean and standard deviation of Max-entropy IRL (Pre-trained).}
\end{table}

\begin{table}[h!]
\centering
\begin{tabular}{c||ccc}
\hline
 & {\textbf{P @ 1}} & {\textbf{P @ 2}} & {\textbf{P @ 3}} \\ \hline\hline
\textbf{GAIL-Mean}    & 0.19                               & 0.452                              & 0.667                              \\
\textbf{GAIL-SD}      & 0.402                              & 0.269                              & 0.211                              \\ \hline
\end{tabular}
\newline
\caption{Mean and standard deviation of GAIL.}
\end{table}

\begin{table}[h!]
\centering
\begin{tabular}{c||ccc}
\hline
                        & \textbf{P @ 1} & \textbf{P @ 2} & \textbf{P @ 3} \\ \hline\hline
\textbf{GAIL(Pre)-Mean} & 0.714          & 0.929          & 0.873          \\
\textbf{GAIL(Pre)-SD}   & 0.463          & 0.179          & 0.166          \\ \hline
\end{tabular}
\newline
\caption{Mean and standard deviation of GAIL (Pre-trained).}
\end{table}

\section*{Appendix E: Omniglot and Multi-digit MNIST}
We recommend reader to learn more about the \href{https://awesomeopensource.com/project/brendenlake/omniglot}{Omniglot} and \href{https://github.com/shaohua0116/MultiDigitMNIST}{Multi-digit MNIST} for few-shot learning tasks. Standard $N$-way-$K$-shot learning tasks often run experiments as $20$-way-$1$-shot, $20$-way-$5$-shot, $5$-way-$1$-shot and $5$-way-$5$-shot. 
\begin{figure}[h!]
    \centering
    \includegraphics[width=0.65\linewidth]{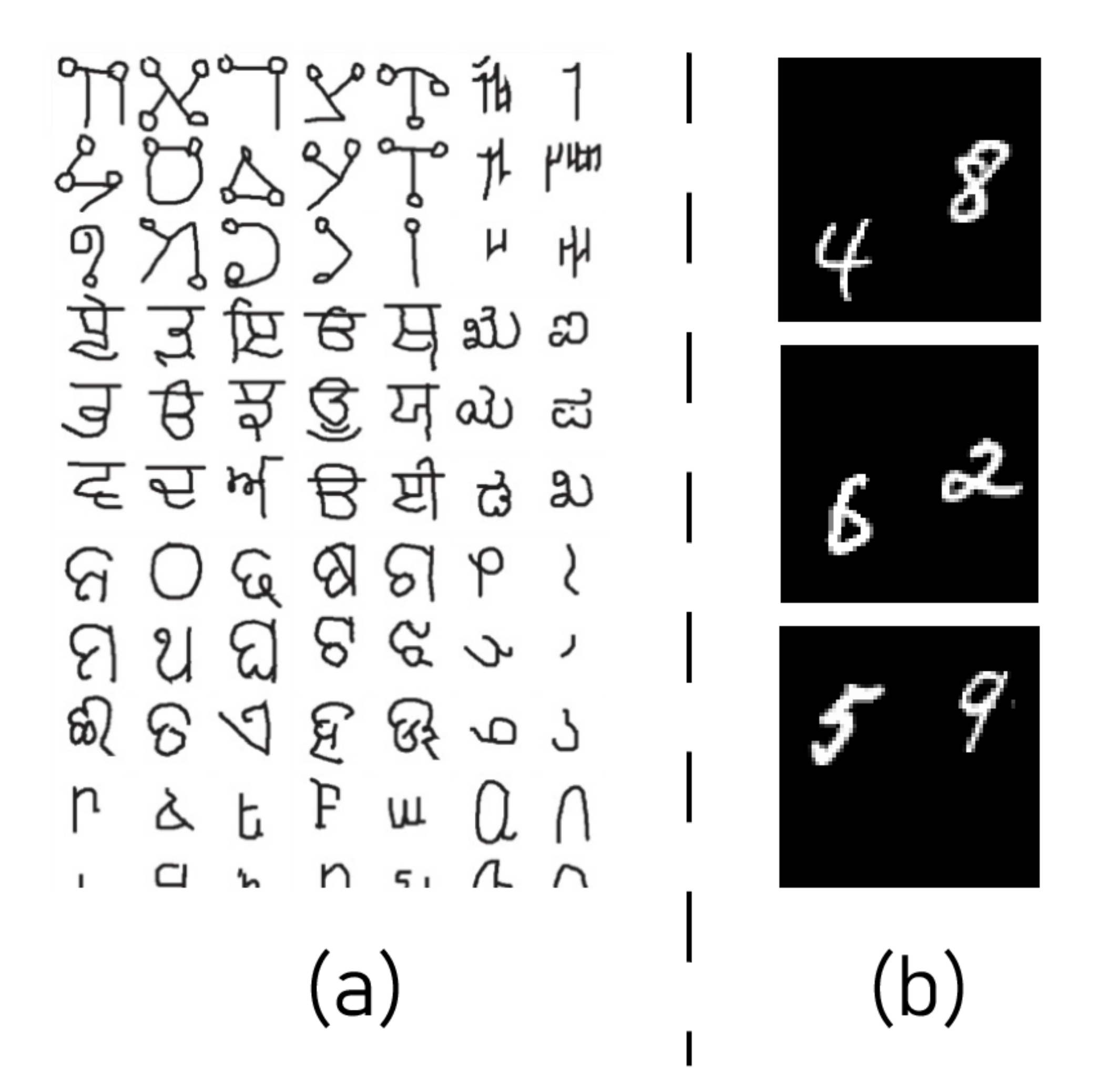}
    \caption{Human handwriting samples from Omniglot and Multi-digit MNIST.}
    \label{fig:my_label}
\end{figure}

The networks used for training MAML, ANIL and PrototypeNet share the same structures for extracting features as the one in Appendix A. They all apply \href{https://arxiv.org/abs/1412.6980}{Adam} as the network optimizer and learning rate is $0.001$.

\end{document}